%% file: acl2023.tex
\pdfoutput=1

\documentclass[11pt]{article}

\usepackage{acl2023}

\usepackage{times}
\usepackage{latexsym}

\usepackage[T1]{fontenc}

\usepackage[utf8]{inputenc}

\usepackage{textcomp}  
\usepackage{csquotes}  

\usepackage{amssymb} 
\usepackage{array}    
\usepackage{adjustbox} 
\usepackage{algpseudocode}
\usepackage{booktabs}
\usepackage{cuted}
\usepackage{enumitem}
\usepackage{float}
\usepackage[linesnumbered,ruled,vlined]{algorithm2e}
\usepackage{listings}
\usepackage{microtype}
\usepackage{multirow}
\usepackage{makecell}
\usepackage[most]{tcolorbox}
\usepackage{pifont}  
\usepackage{siunitx}
\usepackage{tabularx}
\usepackage{threeparttable}
\usepackage{verbatim} 
\usepackage{ragged2e}
\usepackage{wrapfig}
\usepackage{tikz}
\usepackage{subcaption} 
\usetikzlibrary{shapes, arrows.meta, positioning, calc}
\usepackage{url}
\newtcolorbox{taskexample}[1]{
    colback=gray!5, 
    colframe=gray!75, 
    fonttitle=\bfseries,
    title=#1, 
    arc=2mm, 
    boxrule=1pt
}
\usepackage{seqsplit}
\usepackage{fancyvrb}

\lstdefinestyle{json}{
    basicstyle=\small\ttfamily,
    breaklines=true,
    breakatwhitespace=true,
    numbers=left,
    numberstyle=\tiny,
    frame=single,
    showstringspaces=false
}
\usepackage{inconsolata}

\usepackage[dvipsnames, table]{xcolor}

\title{ETOM: A Five-Level Benchmark for Evaluating Tool Orchestration within the MCP Ecosystem}
\author{
  Jia-Kai Dong$^{*,\dagger}$, I-Wei Huang$^{\dagger}$, Chun-Tin Wu, Yi-Tien Tsai \\ 
  National Taiwan University \\
  *\texttt{b11901067@ntu.edu.tw} \\
  $\dagger$ These authors contributed equally to this work
}

\begin{document}

\input{sections/0_abstract}

\input{sections/1_introduction}

\input{sections/2_related_work}

\input{sections/3_msc-bench}

\input{sections/5_benchmark_statistics}

\input{sections/6_experiments_and_results}

\input{sections/7_discussion}

\input{sections/8_limitations}

\input{sections/9_ethical_considerations}

\clearpage


\bibliographystyle{acl_natbib}
\bibliography{anthology,custom}
\clearpage
\appendix
\raggedbottom

\input{sections/A1_algorithm_details}

\input{sections/A2_task_generation_pipelines}

\input{sections/B1_implementation_details}
\input{sections/B2_evaluation_framework}

\input{sections/C1_additional_statistics}

\input{sections/D1_detailed_eval_result}
\input{sections/D2_representative_examples}

\input{sections/E1_prompt_templates}

\end{document}

%% file: sections/0_abstract.tex
\maketitle
\begin{abstract}
We introduce \textbf{ETOM}, a five-level benchmark for evaluating multi-hop, end-to-end tool orchestration by LLM agents within a hierarchical Model-Context Protocol (MCP) ecosystem. Existing benchmarks often assess tools in isolation, overlooking challenges such as functional overlap and cross-server orchestration, which can lead to overly optimistic evaluations. ETOM addresses these gaps by constructing ground truth through ``equal function sets'', enabling objective metrics such as F1 score and reducing reliance on LLM-as-a-judge evaluation. Its five-level curriculum systematically tests agent capabilities, from single-tool orchestration to complex cross-server planning, as well as robustness to out-of-scope requests. Experiments reveal that rigid hierarchies can hinder performance without co-designed strategies, and even state-of-the-art agents exhibit systemic weaknesses in robustness. ETOM provides a diagnostic framework to expose these limitations and guide the development of more capable and efficient tool-using agents.
\end{abstract}

\footnotetext[1]{This paper has been accepted to EACL 2026 Findings.}

%% file: sections/1_introduction.tex
\section{Introduction}
\label{sec:introduction}

\begin{figure*}[t]
    \centering
    \includegraphics[width=0.9\textwidth]{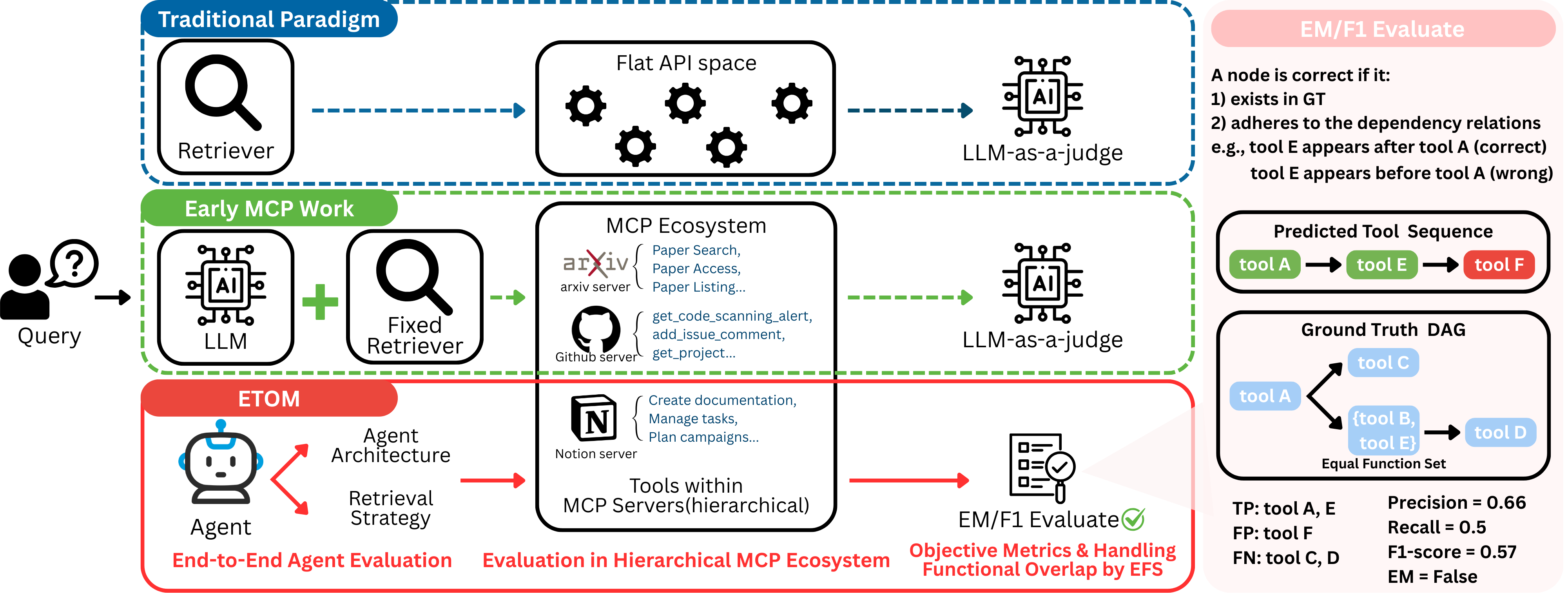}
    \caption{ETOM Overview. The benchmark evaluates end-to-end tool orchestration within a federated MCP ecosystem, featuring 491 servers and 2,375 tools across five complexity levels.}
    \label{fig:overview}
\end{figure*}

\noindent\textbf{The Problem: Evaluating Tool-Using Agents in Real-World Scenarios}

The augmentation of Large Language Models (LLMs) with external tools has become a significant research direction, transforming them from text generators into agents capable of interacting with digital environments~\cite{qin2023toolllmfacilitatinglargelanguage,yao2023reactsynergizingreasoningacting}. A new architectural paradigm is emerging with the Model-Context Protocol (MCP), which organizes tools into semantically coherent, independently operating ``servers''~\cite{anthropicMCP2024}. This federated architecture, analogous to the internet, shifts the agent's task from simply calling an API in a flat namespace to orchestrating workflows across a distributed network. While prior work has shown that a hierarchical organization can improve tool retrieval~\cite{qin2023toolllmfacilitatinglargelanguage, du2024anytoolselfreflectivehierarchicalagents}, the evaluation of end-to-end agent performance in this explicit server-centric reality remains underexplored.

\noindent\textbf{Current Gaps in Existing Benchmarks}

Despite rapid progress, current benchmarks for tool-augmented agents suffer from three fundamental gaps. \textbf{First}, there is an \textbf{architectural mismatch}: most large-scale benchmarks, such as Seal-Tools and NESTful, model tools as a vast, unstructured namespace~\cite{wu2024sealtoolsselfinstructtoollearning, basu2025nestfulbenchmarkevaluatingllms}. This fails to test an agent's ability to navigate the hierarchical, multi-server structure central to the MCP paradigm. For instance, consider a task requiring data from a \textit{Database Server} to be processed by a \textit{Analytics Server}: while both servers may have compatible tools, the agent must understand server boundaries, handle potential connection failures, and orchestrate the workflow across different contexts---challenges that flat-space benchmarks entirely miss. \textbf{Second}, benchmarks struggle with the pervasive challenge of \textbf{functional overlap}, where multiple tools can achieve the same outcome. Some, like ToolHop, meticulously design tasks to avoid this overlap, which limits real-world applicability~\cite{ye2025toolhopquerydrivenbenchmarkevaluating}. Others~\cite{mo2025livemcpbenchagentsnavigateocean, yin2025livemcp101stresstestingdiagnosing, guo2025stabletoolbenchmirrorapimodelingtoolenvironments} rely on an \textit{LLM-as-a-judge} for evaluation, a method known to be costly and susceptible to systemic biases that harm reproducibility~\cite{li2025preferenceleakagecontaminationproblem}. \textbf{Third}, existing evaluation pipelines are often \textbf{fragmented and incomplete}. The modern tool calling system comprises a retriever and an LLM reasoner, but benchmarks tend to evaluate these components in isolation. For example, ToolRet focuses exclusively on retrieval performance~\cite{shi2025retrievalmodelsarenttoolsavvy}, while benchmarks such as NestTools and Seal-Tools evaluate the downstream LLM's reasoning with a fixed, ``golden'' retriever~\cite{han2025nestoolsdatasetevaluatingnested, wu2024sealtoolsselfinstructtoollearning}. This fragmentation provides an incomplete picture of the end-to-end performance of a complete orchestrator.

Table~\ref{tab:benchmark_overview} summarizes these limitations across existing benchmarks, revealing that none provide comprehensive evaluation of both architectural realism (MCP structure) and end-to-end capabilities (LLM+Retriever). Figure~\ref{fig:overview} illustrates how ETOM addresses these fundamental limitations by providing a unified, end-to-end evaluation framework within a realistic MCP ecosystem, contrasting with the fragmented and architecturally misaligned approaches of existing benchmarks.

\noindent\textbf{Our Solution: ETOM}

To address these gaps, we introduce \textbf{ETOM}, a benchmark designed to evaluate end-to-end multihop tool orchestration within an explicit MCP ecosystem with \textbf{491 servers} and \textbf{2375 tools}. Our key innovation is a novel methodology that handles functional overlap by identifying and grouping equivalent tools into \textbf{``equal function sets,''} enabling objective, reproducible evaluation without expensive LLM judges. Building on this foundation, we construct ETOM as a five-level curriculum that systematically ``stress tests'' the full spectrum of agent capabilities:

The curriculum is structured into five levels, starting with foundational single-tool competence (L1-L2), moving to sequential intra-server orchestration (L3), compositional cross-server chaining (L4), and culminating in complex orchestration and robustness against out-of-scope requests (L5).

\begin{figure*}[!t]
    \centering
    \includegraphics[width=\textwidth]{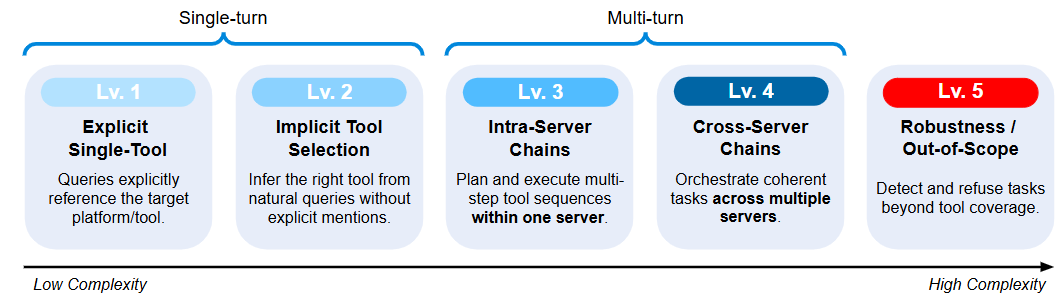}
    \caption{ETOM Five-Level Curriculum Design. The curriculum progresses from foundational single-tool tasks (L1) to complex cross-server orchestration (L4) and robustness testing (L5), systematically evaluating increasingly sophisticated agent capabilities within an MCP ecosystem.}
    \label{fig:five_level_curriculum}
\end{figure*}

Figure~\ref{fig:five_level_curriculum} illustrates the overall structure and progression of our five-level curriculum.

Our extensive experiments on ETOM with a range of agent architectures, from a zero-shot MCP-Zero agent to sophisticated ReAct and retrieval-augmented pipelines, yield critical insights. We find that while current models perform reasonably well in single-tool tasks (L1-L2), their precision degrades significantly in complex multi-server chaining (L4) and robustness checks (L5), often falling below 40\%. These results show that ETOM successfully exposes failure modes in orchestration and robustness that are missed by benchmarks with a narrower scope. Furthermore, our findings challenge the assumption that a hierarchical structure is inherently beneficial. We reveal that without co-designed, hierarchy-aware reasoning strategies, such structures can actually introduce new failure modes and degrade performance. By also measuring latency, we highlight a crucial trade-off between accuracy and efficiency, confirming that an effective orchestrator must balance these competing demands. To facilitate reproducibility, we release the full codebase and dataset at \url{https://github.com/snooow1029/ETOM}.

\noindent\textbf{Contributions}

In summary, our contributions are as follows.
\begin{enumerate}[itemsep=0pt, parsep=0pt, topsep=2pt]
\item A systematic critique of the existing evaluation landscape for tool-using agents, identifying fundamental gaps in architectural alignment, evaluation methodology, and scope.
\item A novel methodology based on ``equal function sets'' enables objective, reproducible, and efficient evaluation of tool orchestrations in the presence of functional overlap.
\item The release of \textbf{ETOM}, the first large-scale, five-level benchmark to evaluate end-to-end, multihop tool orchestration within a multi-server MCP ecosystem.
\item A comprehensive experimental analysis that reveals key weaknesses of current agent architectures regarding cross-server orchestration, robustness, and the trade-offs between hierarchy, accuracy, and latency.
\end{enumerate}

%% file: sections/2_related_work.tex
\section{Related Work}
\label{sec:related_work}

The landscape of tool-augmented agent benchmarks has rapidly evolved from assessing foundational skills to complex agentic behaviors~\cite{ferrag2025llmreasoningautonomousai}. We position ETOM by examining prior work across three key themes: flat-namespace and capability-specific benchmarks, broader agentic benchmarks, and the emerging hierarchical MCP-aligned ecosystem.

\paragraph{Flat-Namespace and Capability-Specific Benchmarks.}
Prior large scale benchmarks ~\cite{qin2023toolllmfacilitatinglargelanguage, wu2024sealtoolsselfinstructtoollearning} model tools as a monolithic, flat namespace, failing to test multi-server navigation. Subsequent works target specific capabilities: ComplexFuncBench~\cite{zhong2025complexfuncbenchexploringmultistepconstrained} and BFCL v2~\cite{mao2024bfcl} focus on complex function-calling, while ToolHop~\cite{ye2025toolhopquerydrivenbenchmarkevaluating} assesses multi-hop reasoning by avoiding functional overlap. These benchmarks evaluate components in isolation, overlooking cascading errors in end-to-end systems.

\paragraph{Agentic and Interactive Benchmarks.}
Recent benchmarks evaluate holistic agentic capabilities in realistic settings: GAIA~\cite{mialon2023gaiabenchmarkgeneralai} for problem-solving, SWE-Lancer~\cite{miserendino2025swelancerfrontierllmsearn} for software engineering, and MultiAgentBench~\cite{zhu2025multiagentbenchevaluatingcollaborationcompetition} for multi-agent collaboration. However, these rely on LLM-as-a-judge evaluation, which suffers from systemic biases and low reproducibility~\cite{tan2025judgebenchbenchmarkevaluatingllmbased, zhuge2024agentasajudgeevaluateagentsagents}.

\paragraph{Hierarchical and MCP-Aligned Benchmarks.}
The Model Context Protocol (MCP)~\cite{anthropicMCP2024} offers a standardized, federated architecture. Nascent benchmarks like MCP-RADAR~\cite{gao2025mcpradarmultidimensionalbenchmarkevaluating} and LiveMCPBench~\cite{mo2025livemcpbenchagentsnavigateocean} adopt this structure but rely on fallible LLM judges and sidestep \textbf{functional overlap}.

ETOM addresses these gaps by providing architectural realism through a large-scale MCP ecosystem, comprehensive end-to-end evaluation, and a novel ``equal function sets'' methodology enabling objective metrics even with functional overlap. Table~\ref{tab:benchmark_overview} provides quantitative comparison.

\newcommand{\cmark}{\ding{51}} 
\newcommand{\xmark}{\ding{55}} 

\begin{table*}[!t]
\centering
\tiny
\begin{adjustbox}{width=\textwidth}
\begin{tabular}{l c c c c c c}
\toprule
\textbf{Benchmark} & \textbf{Tools (S/T)} & \textbf{Source} & \textbf{Tasks} & \textbf{Capabilities} & \textbf{Target} & \textbf{Eval} \\
\midrule
\multicolumn{7}{c}{\textit{MCP Benchmarks}} \\
\midrule
MCP-RADAR~\cite{gao2025mcpradarmultidimensionalbenchmarkevaluating} & 9/42 & Real. & 300 & ST+MT & L(all in prompt) & Rule \\
MCPEval~\cite{liu2025mcpevalautomaticmcpbaseddeep} & 12/77 & Real. & 676 & N/A & L(MCP interface) & Rule+LLM \\
LiveMCPBench~\cite{mo2025livemcpbenchagentsnavigateocean} & 70/527 & Real. & 95 & MT & L(MCP-Zero) & LLM \\
MCP-Universe~\cite{luo2025mcpuniversebenchmarkinglargelanguage} & 11/133 & Real. & 231 & MT & L(oracle R) & LLM \\
\textbf{ETOM (ours)} & \textbf{491/2375} & \textbf{Real.} & \textbf{2075} & \textbf{ST+MT+R} & \textbf{L+R} & \textbf{Rule} \\
\midrule
\multicolumn{7}{c}{\textit{Tool Benchmarks}} \\
\midrule
NestTools~\cite{han2025nestoolsdatasetevaluatingnested} & N/A/3034 & Syn. & 1000 & MT & L(gte-large) & Rule \\
Seal-Tools~\cite{wu2024sealtoolsselfinstructtoollearning} & N/A/4076 & Syn. & 14076 & ST+MT & L(DPR) & Rule \\
ToolRet~\cite{shi2025retrievalmodelsarenttoolsavvy} & N/A/44k & Mix. & 7961 & ST+MT & R & Rule \\
\bottomrule
\end{tabular}
\end{adjustbox}
\caption{Benchmark comparison. S/T = Servers/Tools. ST=single-tool, MT=multi-tool, R=robustness. L = evaluated only with LLM, R = evaluated only with Retriever. Parentheses indicate Retriever used during LLM evaluation.}
\label{tab:benchmark_overview}
\end{table*}

%% file: sections/3_msc-bench.tex
\section{ETOM: Design and Construction}
\label{sec:benchmark_construction}

ETOM is a large-scale benchmark designed for the hierarchical MCP ecosystem, systematically addressing the challenge of function overlap that pervades real-world tool orchestration. Constructing such a benchmark presents significant challenges: it must be large-scale, realistic, and capable of evaluating a wide spectrum of agentic abilities while accounting for the inherent functional redundancy in hierarchical tool systems. ETOM addresses these systematically through four core stages: 
(1) building a comprehensive, real-world MCP tool corpus; 
(2) identifying functionally equivalent tools to resolve function overlap; 
(3) generating a five-level task curriculum that leverages the hierarchical nature of the ecosystem; 
and (4) ensuring data quality and validity across all complexity levels.

Full procedural details for each stage, including the tool filtering criteria, the equal function set generation algorithm, and task generation pipelines for all five levels, are provided in Appendix~\ref{app:server_filtering_prompt}. Representative examples for each task level can also be found in Appendix~\ref{app:representative_examples}.

\subsection{Corpus Construction: A Real-World MCP Toolset}
\label{subsec:corpus_construction}

The foundation of ETOM is a diverse tool corpus sourced from the \texttt{glama.ai} MCP server registry~\cite{glama_ai}. We scraped the top 50 servers from each category and performed a rigorous filtering process to ensure the quality and relevance of the toolset. Inspired by the guidelines for identifying confounding tools in benchmarks~\cite{huang2024metatoolbenchmarklargelanguage}, we first conducted a semi-automated filtering process to exclude servers that do not represent genuine, indispensable external capabilities. 

This process targeted several categories of unsuitable servers: 
\begin{enumerate}[itemsep=0pt, topsep=2pt]
    \item \textbf{Trivial Tools}, whose functions are subsumed by the native capabilities of modern LLMs (e.g., simple calculators);
    \item \textbf{Meta-Tools}, which are designed to augment the agent's own internal processes (e.g., memory, reasoning patterns); 
    \item \textbf{Template Servers}, which primarily serve as developer examples and lack a cohesive use case.
\end{enumerate}

Each server was evaluated against a strict definition of ``native LLM capability''---defined as tasks a pure, sandboxed LLM could perform without any external tools. The full methodology, including the detailed criteria and the prompt used for our LLM-based analysis, is provided in Appendix~\ref{app:server_filtering_prompt}. This multi-stage process, combining automated analysis with human review, ensured that the final toolset is composed of meaningful and externally-focused tools suitable for evaluating complex orchestration.

The resulting filtered corpus contains \textbf{491 unique servers} exposing \textbf{2,375 distinct tools}, each represented by a JSON object with its name, description, and formal input schema (see Appendix~\ref{app:mcp_data_format}). This hierarchical corpus establishes a realistic and challenging environment for all benchmark tasks.

\subsection{Identifying Functional Overlap: The Equal Function Set Methodology}
\label{subsec:equal_function_sets}

Functional overlap complicates evaluation, as multiple tools can fulfill the same user intent. We solve functional overlap through a round-trip consistency approach that integrates \textbf{bottom-up} and \textbf{top-down} verification to ensure comprehensive and practically relevant equal function sets. The integration of these processes establishes \emph{round-trip consistency}, ensuring functional equivalence is coherent at both the corpus level and within actual query contexts. This ensures all candidate tools associated with a Level 2 query are functionally equivalent, forming robust ground truth for evaluation. Furthermore, these equal function sets serve as the cornerstone for Level 4 cross-server compositional tasks, enabling systematic identification of functionally equivalent tools across different servers. Complete algorithmic details are provided in Appendix~\ref{app:algorithm_details}.

\paragraph{Bottom-up: Candidate Generation and Pairwise LLM Verification}
For each tool, we retrieve semantically similar tools using Qwen3-Embedding-0.6B (similarity $> 0.8$). A large LLM performs pairwise verification to determine functional equivalence, with verified pairs grouped into connected components using Union--Find to form \emph{equal function sets}. This stage identifies 145 initial candidate sets based on semantic descriptions.

\paragraph{Top-down: Query-Guided RAG and Human Verification}
To ensure equivalence in realistic usage, each Level 2 query retrieves the top-10 relevant tools via RAG. An LLM selects all tools capable of fulfilling the query. These associations are cross-checked against the bottom-up sets. Finally, human validators manually examine borderline cases and platform-specific nuances (e.g., ensuring consistency across different cloud providers). This rigorous two-stage process prunes the candidates into 95 high-confidence equal function sets, ensuring logical and platform-level consistency for the evaluation ground truth.

\subsection{Task Generation: A Five-Level Curriculum}
\label{subsec:task_generation}

ETOM implements a five-level curriculum designed to systematically evaluate agent capabilities (see Figure~\ref{fig:five_level_curriculum}). The levels progress from foundational single-tool tasks (L1-L2) to complex, multi-hop orchestration (L3-L4) and robustness testing (L5). Each level targets a distinct reasoning challenge. 

\paragraph{Tool Triage and Semantic Annotation}
Prior to task generation, all tools undergo annotation along three axes: \textit{Platform Specificity} (distinct platform requirements), \textit{Task Type} (final-goal vs. middleware), and \textit{User Orientation} (user-facing vs. system-facing). Only tools meeting all criteria are considered for Level 1 and Level 2 generation.

\noindent\textbf{Level 1: Foundational Single-Tool Tasks}
These tasks establish baseline competence through direct tool invocation. A Generator-Verifier pipeline produces queries that satisfy semantic and structural constraints, ensuring each task has a clear, unambiguous solution path.

\noindent\textbf{Level 2: Context-Aware Tool Retrieval}
This level tests disambiguation capabilities when multiple tools could potentially fulfill user intent. Preliminary L2 queries validate and extend equal function sets via RAG retrieval and LLM verification, ensuring the dataset is diverse and grounded in validated tool equivalences.

\noindent\textbf{Level 3: Intra-Server Sequential Chaining}
Complex workflows within individual servers require understanding tool dependencies and data flow. The pipeline infers output schemas using an LLM, constructs data-flow dependency graphs, and identifies valid chains where outputs satisfy subsequent inputs, generating tasks with Chain-of-Thought rationales.

\noindent\textbf{Level 4: Cross-Server Compositional Chaining}
Multi-server orchestration demands reasoning across different contexts and handling potential failures. Two to four servers are sampled from distinct categories, and a planner LLM constructs coherent cross-server task flows. Generated tasks are verified for realism and executability.

\noindent\textbf{Level 5: Robustness via Capability Gap Identification}
Agents must recognize when requests exceed their capabilities rather than attempting impossible tasks. The pipeline clusters tool embeddings to map capabilities, identifies gaps where no existing tool can fully address tasks, generates natural language queries for each gap, and validates that queries cannot be solved with the current toolset.

\begin{table}[!htbp]
    \centering
    \resizebox{\columnwidth}{!}{%
    \begin{tabular}{lrrr} 
        \toprule
        \thead{Level and Key Challenge} & \thead{\# of Tasks} & \thead{Avg. Plan Length} & \thead{Avg. Servers} \\
        \midrule
        \makecell[l]{L1: Direct Retrieval \\ \textit{\small Foundational: Direct Tool Retrieval}}         & 781   & 1.00 & 1.00 \\
        \addlinespace 
        \makecell[l]{L2: Context-Aware Retrieval \\ \textit{\small Disambiguation among Functional Overlap}}  & 773   & 1.00 & 1.00 \\
        \addlinespace
        \makecell[l]{L3: Intra-Server Chaining \\ \textit{\small Orchestration: Intra-Server Sequential Chaining}}    & 327   & 2.87 & 1.00 \\
        \addlinespace
        \makecell[l]{L4: Cross-Server Chaining \\ \textit{\small Orchestration: Cross-Server Compositional Chaining}}    & 103   & 3.83 & 3.78 \\
        \addlinespace
        \makecell[l]{L5: Robust Rejection \\ \textit{\small Robustness: Rejection of Out-of-Scope Requests}}         & 91    & 0.00 & 0.00 \\
        \bottomrule
    \end{tabular}%
    }
    \caption{Overview of the ETOM evaluation curriculum statistics.}
    \label{tab:curriculum_summary}
\end{table}

\subsection{Data Quality and Validation}
All levels incorporate a multi-stage verification pipeline to ensure tasks are pragmatically plausible, executable, and internally consistent.
\textbf{Automated and Structural Validation:} For all levels, a verifier LLM validates the solvability of each task. For multi-step tasks (L3--L4), we implement \emph{pseudo output schemas}—including name, description, inputs, and outputs—to ensure subsequent steps are grounded in realistic data dependencies. This design preserves the logical fidelity of multi-hop reasoning while maintaining benchmark accessibility without requiring private API keys.
\textbf{Human-in-the-Loop Quality Control:} We conducted manual inspections on a random sample of 100 data entries from all five levels. Human annotators verified that queries were natural, unambiguous, and that the ground truth tool sequences correctly addressed the user intent. This dual validation (LLM + Human) ensures high-quality tasks that accurately reflect real-world orchestration challenges. Complete procedural details and validation statistics are provided in Appendix~\ref{app:task_generation_pipelines}.

%% file: sections/5_benchmark_statistics.tex
\begin{table*}[!t]
\centering
\resizebox{\textwidth}{!}{%
\begin{tabular}{llcccccccccc}
\toprule
\multirow{2}{*}{\textbf{Architecture}} & \multirow{2}{*}{\textbf{Foundation Model}} & \multicolumn{2}{c}{\textbf{Level 1}} & \multicolumn{2}{c}{\textbf{Level 2}} & \multicolumn{2}{c}{\textbf{Level 3}} & \multicolumn{2}{c}{\textbf{Level 4}} & \multicolumn{2}{c}{\textbf{Level 5}} \\
\cmidrule(lr){3-4} \cmidrule(lr){5-6} \cmidrule(lr){7-8} \cmidrule(lr){9-10} \cmidrule(lr){11-12}
& & \textbf{EM (\%)} & \textbf{N-Lat.} & \textbf{EM (\%)} & \textbf{N-Lat.} & \textbf{F1 (\%)} & \textbf{N-Lat.} & \textbf{F1 (\%)} & \textbf{N-Lat.} & \textbf{EM (\%)} & \textbf{N-Lat.} \\
\midrule
\multicolumn{12}{c}{\textit{MCP-Zero (MCP0)}} \\
\midrule
MCP0 & Qwen3-4B-Instruct-2507 & 69.01 & \textbf{0.89} & 43.02 & \textbf{0.59} & 46.85 & 5.64 & 30.42 & 6.50 & 74.72 & \textbf{0.22} \\
MCP0 & Qwen3-8B & 65.68 & 3.14 & 31.56 & 3.60 & 48.16 & 3.87 & 33.95 & 3.73 & 62.63 & 3.86 \\
MCP0 & Ministral-8B-Instruct-2410 & 46.99 & 1.29 & 23.67 & 1.24 & 39.65 & 1.43 & 25.99 & 1.71 & 51.60 & 0.70 \\
MCP0 & GPT-4.1 & 63.38 & \textit{--} & 31.82 & \textit{--} & 53.12 & \textit{--} & 34.20 & \textit{--} & 80.29 & \textit{--} \\
MCP0 & Gemma-3-12B-IT & 70.55 & 1.73 & 33.50 & 1.56 & 49.70 & 1.71 & 32.43 & 1.99 & 65.93 & 0.75 \\
MCP0 & Microsoft Phi-4 & 38.79 & 3.54 & 19.01 & 3.37 & 53.97 & 3.31 & 33.98 & 3.42 & 69.23 & 2.58 \\
MCP0 & Meta-Llama-3-8B-Instruct & 48.08 & 1.00 & 19.25 & 1.00 & 38.04 & \textbf{1.00} & 25.70 & \textbf{1.00} & 39.56 & 1.00 \\
\midrule
\multicolumn{12}{c}{\textit{ToolShed (TS)}} \\
\midrule
TS & Qwen3-4B-Instruct-2507 & 72.72 & 3.31 & 51.87 & 3.58 & 69.41 & 5.38 & 37.45 & 5.26 & 70.32 & 2.22 \\
TS & Qwen3-8B & 80.65 & 11.63 & 53.29 & 11.33 & 77.45 & 15.74 & 44.96 & 15.31 & 76.92 & 11.25 \\
TS & Ministral-8B-Instruct-2410 & 64.66 & 7.86 & 43.46 & 8.24 & 82.36 & 10.15 & 50.98 & 8.68 & 45.05 & 4.70 \\
TS & GPT-4.1 & 69.91 & \textit{--} & 49.02 & \textit{--} & \textbf{82.55} & \textit{--} & \textbf{55.06} & \textit{--} & \textbf{81.31} & \textit{--} \\
TS & Gemma-3-12B-IT & \textbf{80.66} & 7.89 & \textbf{56.79} & 7.87 & 73.96 & 11.80 & 47.29 & 12.19 & 60.43 & 6.07 \\
TS & Microsoft Phi-4 & 38.66 & 15.12 & 28.71 & 14.54 & 81.85 & 14.52 & 49.67 & 14.81 & 67.03 & 11.93 \\
TS & Meta-Llama-3-8B-Instruct & 57.03 & 5.22 & 42.12 & 6.02 & 74.34 & 5.22 & 50.77 & 5.68 & 27.40 & 6.65 \\
\midrule
\multicolumn{12}{c}{\textit{ReAct}} \\
\midrule
ReAct & Qwen3-4B-Instruct-2507 & 41.43 & 8.80 & 19.76 & 8.65 & 33.61 & 6.75 & 6.74 & 10.54 & 72.50 & 12.18 \\
\midrule
\multicolumn{12}{c}{\textit{Hybrid}} \\
\midrule
Hybrid & Qwen3-4B-Instruct-2507 & 68.62 & 3.61 & 60.20 & 3.68 & 56.91 & 3.63 & 26.78 & 2.37 & 74.72 & 1.58 \\
Hybrid & Meta-Llama-3-8B-Instruct & 48.08 & 4.74 & 39.79 & 4.39 & 50.89 & 6.07 & 21.61 & 5.48 & 14.28 & 4.94 \\
\bottomrule
\end{tabular}
}
\caption{Detailed Performance and Latency Results. Performance for L1, L2, and L5 is measured by Exact Match (EM), while L3 and L4 use F1-score. Normalized Latency (N-Lat.) is relative to the MCP0 Meta-Llama-3-8B-Instruct baseline. The best performance in each column is \textbf{bolded}. \textit{Latency values for API-based models (e.g., GPT-4.1) are omitted (--) because they are not directly comparable to locally deployed models due to network overhead and hardware variability.
}}
\label{tab:appendix_full_results}
\end{table*}

\section{Benchmark Statistics}
\label{sec:benchmark_statistics}

We present key statistics of ETOM, highlighting its scale and comprehensive five-level evaluation curriculum (see Figure~\ref{fig:five_level_curriculum} for the curriculum design). An overview of statistics for each level is provided in Table~\ref{tab:curriculum_summary}.

\noindent\textbf{Comparative Analysis of Retrieval Complexity}
\label{subsec:complexity_analysis}
We analyze our foundational retrieval tasks (L1 and L2) against prominent single-step benchmarks using lexical (ROUGE-L) and semantic (METEOR) similarity metrics. Our analysis, detailed in \textbf{Table~\ref{tab:complexity_comparison_appendix}} in the Appendix, confirms that tool retrieval in existing benchmarks relies on semantic understanding (low lexical overlap). ETOM contributes to this paradigm with its low-overlap L2 tasks, while also filling a gap with its high-overlap L1 tasks that test direct command following. This dual approach provides comprehensive evaluation coverage. Note that multi-step tasks (L3-L4) require graph-based metrics rather than lexical similarity measures.

\noindent\textbf{Level 5: Robustness via Capability Gap }

Level 5 comprises 91 out-of-scope queries targeting \textbf{31 distinct capability gaps} identified through systematic analysis. These gaps represent well-defined categories (e.g., physical world interaction, real-time sensory processing) outside the digital MCP ecosystem boundaries. The distribution of these categories is shown in Appendix~\ref{app:level5_dist}, ensuring comprehensive robustness testing.


%% file: sections/6_experiments_and_results.tex
\begin{table}[H]
\centering
\resizebox{\linewidth}{!}{
\begin{tabular}{llc}
\toprule
Level & Objective & Metric \\
\midrule
L1 & Direct tool retrieval & Exact Match (EM) \\
L2 & Context-Aware Tool Retrieval & EM against set \\
L3 & Intra-server task orchestration & Node Set EM, F1 \\
L4 & Cross-server task orchestration & Node Set EM, F1 \\
L5 & Reject unfulfillable requests & Exact Rejection Match (ERM) \\
\bottomrule
\end{tabular}
}
\caption{Evaluation Levels and Metrics Summary}
\label{tab:eval_summary}
\vspace{-1em}
\end{table}

\section{Experiments and Results}
\label{sec:experiments}

We evaluate four tool orchestration frameworks on ETOM, examining performance across complexity levels, efficiency trade-offs, and key hyperparameter effects.

\subsection{Experimental Setup}
\begin{figure}[H]
\centering
\includegraphics[width=0.9\linewidth]{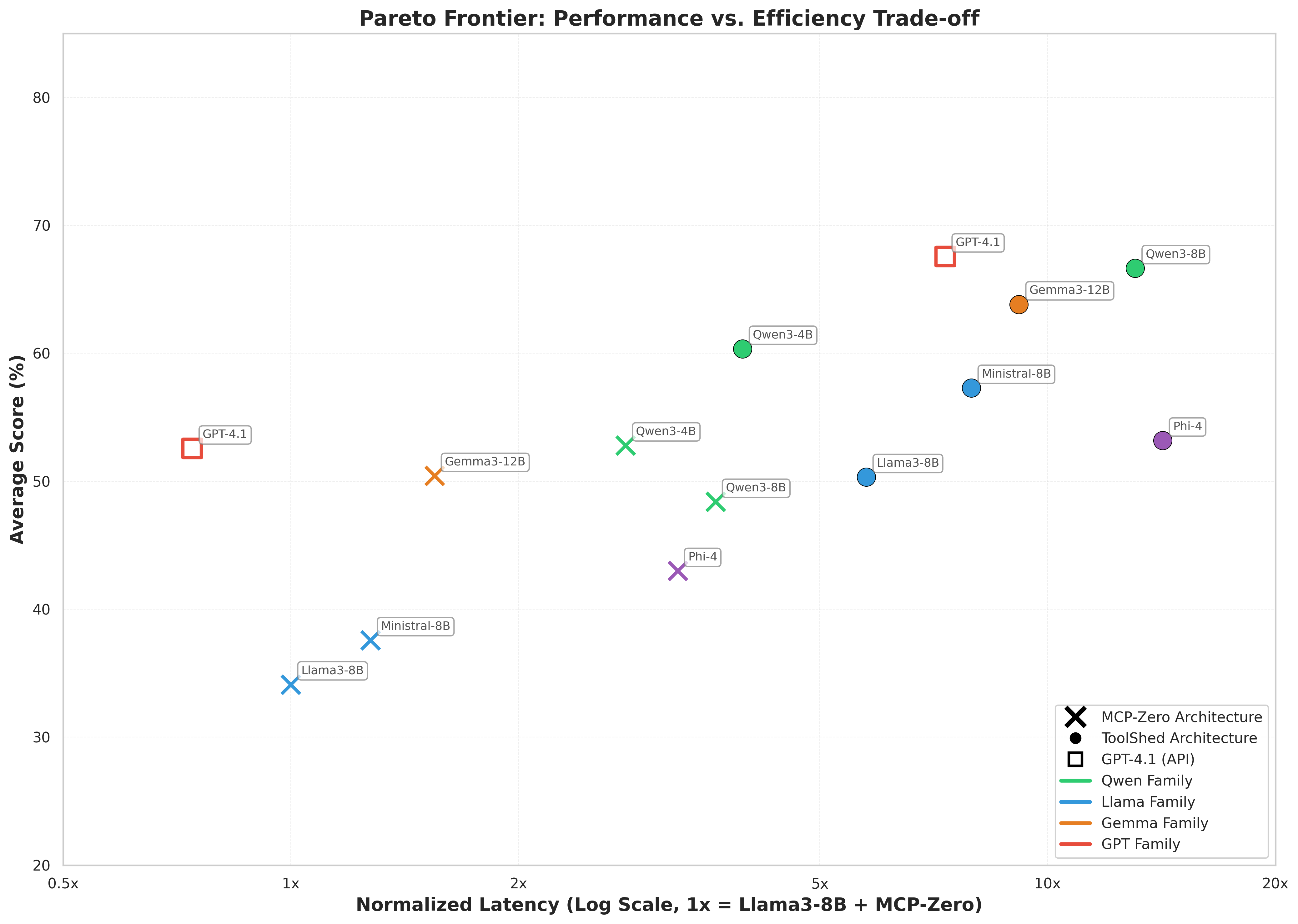}
\caption{Performance vs. efficiency across architectures and foundation models. Each point is a model–architecture combination; markers: X = MCP-Zero, circle = ToolShed. The x-axis uses a log scale for normalized latency.}
\label{fig:performance_efficiency_tradeoff}
\vspace{-1em}
\end{figure}

Four architectures: \textbf{ReAct}~\cite{yao2023reactsynergizingreasoningacting} (generative baseline), \textbf{ToolShed}~\cite{lumer2024toolshedscaletoolequippedagents} (flat retrieval with dense search, query expansion, re-ranking), \textbf{MCP-Zero}~\cite{fei2025mcpzeroactivetooldiscovery} (hierarchical retrieval), and \textbf{Hybrid} (combining MCP-Zero filtering with ToolShed search). Foundation models: \textbf{Qwen3-4B-Instruct}, \textbf{Qwen3-8B}~\cite{yang2025qwen3technicalreport}, \textbf{Meta-Llama-3-8B-Instruct}~\cite{grattafiori2024llama3herdmodels}, \textbf{Ministral-8B-Instruct-2410}~\cite{mistral2024ministral8b}, \textbf{GPT-4.1}~\cite{openai2024gpt4technicalreport}, \textbf{Gemma-3-12B-IT}~\cite{gemmateam2025gemma3technicalreport}, and \textbf{Microsoft Phi-4}~\cite{abdin2024phi4technicalreport}. We evaluate this comprehensive set of models to assess backbone sensitivity and model-architecture interactions; aggregate trends appear in Figure~\ref{fig:performance_efficiency_tradeoff}, with main experimental results in Table~\ref{tab:appendix_full_results} and ablation study of parameters in each architecture in Appendix~\ref{app:detailed_evaluation_results}.

\paragraph{Evaluation Methodology}
We evaluate the proposed orchestrator agent using a multi-level framework (L1--L5) designed to probe distinct facets of tool-use reasoning, ranging from foundational retrieval to multi-step task decomposition and robust rejection of infeasible requests (see Figure~\ref{fig:five_level_curriculum}). To ensure fair cross-level comparison, all evaluations employ a \textbf{fixed agent architecture}, isolating observed performance differences to reasoning capability rather than ad-hoc modifications.

Table~\ref{tab:eval_summary} provides a high-level overview of each evaluation level with its primary objective and corresponding metric. Full procedural details, including scoring rules and verification protocols, are available in Appendix~\ref{app:evaluation_framework}.

\subsection{Main Performance Comparison}
Our results in Table~\ref{tab:appendix_full_results} reveal a clear performance hierarchy. First, \textbf{retrieval-augmented frameworks decisively outperform the generative baseline}, with ToolShed-Qwen-4B-Instruct (60.35) nearly doubling the score of ReAct-Qwen-4B-Instruct (34.81). This confirms that direct access to a tool repository is superior to relying on a model's parametric memory.
Among retrieval methods, \textbf{ToolShed's flat-retrieval architecture achieves the highest overall scores}. Its comprehensive search and re-ranking prove particularly effective for complex orchestration, where Llama-ToolShed excels on L3 (74.34) and L4 (50.77) tasks. However, performance is not determined by architecture alone. We observe a strong model dependency: Qwen demonstrates strength in direct retrieval (L1-L2), whereas Llama's advanced reasoning shines in multi-step tasks (L3-L4). This is most evident in our Hybrid architecture, where performance with Qwen is competitive (57.45), but drops dramatically with Llama (34.93).

\subsection{Efficiency Analysis}

Table~\ref{tab:appendix_full_results} shows normalized latency for our locally-run configurations, while Figure~\ref{fig:performance_efficiency_tradeoff} visualizes the performance-efficiency trade-off.

Our analysis reveals a clear architectural clustering and a fundamental trade-off. Configurations using our \textbf{MCP-Zero} architecture (marked with 'x') consistently occupy the \textbf{high-efficiency, low-latency region} (1.0-3.0x baseline). In contrast, \textbf{ToolShed} configurations (marked with \textbullet) cluster in the \textbf{high-performance, high-latency region} (5.0-15.0x). This demonstrates that hierarchical filtering buys efficiency by sacrificing comprehensiveness, while flat retrieval purchases performance at a steep computational cost.

The choice of foundation model further defines the Pareto frontier within these clusters. For our locally-run models:
\begin{itemize}[itemsep=2pt, parsep=0.5pt, topsep=2pt]
    \item Under \textbf{MCP-Zero}, \textbf{Qwen} variants sit on the efficient frontier, achieving the lowest normalized latency on single-step tasks (L2/L5).
    \item Under \textbf{ToolShed}, stronger backbones like \textbf{Gemma3-12B} dominate the Pareto frontier, pushing the performance boundaries for locally-run models.
\end{itemize}

The API-based \textbf{GPT-4.1} (marked with a distinct square $\Box$ in Figure~\ref{fig:performance_efficiency_tradeoff}) serves as a crucial, high-performance reference point. While its latency is not directly comparable to our local models, its performance positions it as the anchor of the ``performance frontier,'' particularly on complex L3/L4 tasks. The ToolShed+GPT-4.1 configuration defines the current SOTA in terms of raw capability, qualitatively reinforcing the high-performance, high-latency nature of the ToolShed architecture. The full implications of this trade-off are explored in Section~\ref{sec:discussion}.

\subsection{Ablation Study of ToolShed’s Architecture}
\label{sec:ablation}
To understand the impact of different components in the ToolShed architecture, we conducted a thorough ablation study. Our key findings indicate that retrieval breadth (the number of retrieved tools) has a significant, task-specific impact: broader retrieval helps complex L3/L4 tasks but can degrade performance on simpler L2 tasks due to noise. We also found that query expansion offers context-dependent benefits, while reranking is critical for selection quality across all task levels.

\textbf{Due to space limitations, we provide the complete analysis in Appendix~\ref{app:ablation}, while the main text summarizes the essential findings.}

\subsection{Failure Analysis and Complexity Decay}
\label{sec:failure-analysis}

To diagnose agent limitations, we conduct a qualitative review of failure traces in Levels~3 and~4. Our analysis identifies two primary failure patterns that hinder complex orchestration:

\paragraph{Catastrophic Context Loss in Multi-hop Planning}
The most prevalent failure in Level~4 arises when agents are required to orchestrate tools across multiple servers. We observe that agents frequently lose critical context (e.g., intermediate tool outputs or explicit user constraints) when transitioning across server boundaries. This phenomenon, which we term \emph{contextual drift}, triggers a cascade of errors in which subsequent tool selections become increasingly decoupled from the original user intent.

\paragraph{Premature Decomposition}
Many agents exhibit a tendency to decompose simple, single-step queries into unnecessary sub-tasks. This over-decomposition often introduces hallucinated dependencies that are not present in the tool documentation, thereby increasing the complexity of the execution plan and creating additional failure points in the reasoning chain.

\paragraph{Quantitative Complexity Decay}
These qualitative failure modes are reflected in a pronounced performance degradation as task length increases. Analysis of Level~3 (Sequential Tasks) reveals a consistent trend: for the ToolShed-Qwen-4B configuration, performance drops from 70.2\% F1 on 2-step tasks to 43.1\% F1 on 7-step tasks. This degradation pattern remains stable across different foundation models, suggesting that maintaining long-horizon reasoning constitutes a systemic challenge in federated MCP environments.

%% file: sections/7_discussion.tex
\section{Discussion}
\label{sec:discussion}

Building on the experimental results, this section delves into their broader implications. Our empirical evaluation reveals fundamental insights about tool orchestration systems through three critical lenses: the hidden costs of architectural choices, the intricate dance between models and architectures, and the most persistent challenges that remain unsolved.

\subsection{Hierarchy Isn't Free: The Hidden Costs of Architectural Choices}
Our results challenge the assumption that hierarchical retrieval is inherently superior. While MCP-Zero's architecture delivers impressive efficiency gains (up to 5.76x), this benefit comes at a significant and non-uniform accuracy cost, particularly in complex orchestration tasks (L3/L4) where its rigid constraints appear to be limiting. This suggests a provocative possibility: that rigid, pre-defined hierarchies, while computationally attractive, are fundamentally antithetical to the fluid, associative reasoning style of modern LLMs. By forcing a model's reasoning into a fixed tree structure, we may be inadvertently stifling the very capabilities we seek to leverage.

Our experimental \textbf{Hybrid} model further exposes this tension—the dramatic performance drop with Llama (34.93 vs 50.33 for ToolShed) demonstrates that simply combining paradigms is insufficient. True hybrid systems require architectural innovation that goes beyond straightforward fusion, potentially involving dynamic switching or context-aware retrieval strategies.

\subsection{Model-Architecture Co-Design: Why There's No Universal Best}

A striking finding from our experiments is that model performance is not an intrinsic property but an emergent one, arising from the complex interplay between a foundation model's reasoning style and its orchestration architecture. This highlights a critical challenge in \textbf{model-architecture co-design}.

This tension is starkly illustrated by our experimental \textbf{Hybrid} architecture. Its dramatic performance drop when switching from Qwen to Llama (from a competitive 57.45 to a poor 34.93) reveals that simply fusing two powerful paradigms (hierarchical filtering and flat retrieval) is insufficient. The failure suggests a deep incompatibility, or ''impedance mismatch,'' between Llama's reasoning patterns and the specific architectural constraints of the Hybrid model, proving that architectural benefits are not universally transferable across all models.

More broadly, our results show that different models exhibit complementary strengths which are either amplified or constrained by the architectural framework:
\begin{itemize}[itemsep=2pt, parsep=0.5pt, topsep=2pt]
    \item \textbf{Qwen} models demonstrate superior precision in direct, single-step retrieval tasks (L1/L2). This suggests they excel when the search space is well-defined and the task requires accurate instruction-following, a strength that benefits from MCP-Zero's efficient pruning.
    \item In contrast, \textbf{Llama}'s advanced reasoning capabilities for complex, multi-step tasks (L3/L4) only fully manifest when paired with ToolShed's broad, high-recall retrieval mechanism. This indicates that sophisticated reasoning requires a comprehensive set of candidate tools to operate on, a condition that a restrictive hierarchy may fail to provide.
\end{itemize}

This symbiotic relationship has profound implications: there is no ``best'' model or ``best'' architecture in isolation, only \textbf{optimal model-architecture pairs} for specific task profiles. Future systems must therefore treat model selection and architectural design not as separate optimization problems, but as a unified co-design challenge.

\paragraph{Link to performance-efficiency trade-offs}
Figure~\ref{fig:performance_efficiency_tradeoff} makes these interactions visible: \emph{Qwen+MCP-Zero} configurations populate the efficient frontier for simpler workloads (L1/L2), while \emph{Llama+ToolShed} configurations occupy the performance frontier for complex orchestration (L3/L4) at higher latency. Stronger backbones (e.g., GPT-4.1, Gemma3-12B) extend the frontier under ToolShed without collapsing efficiency under MCP-Zero, underscoring that backbone choice and retrieval architecture must be co-optimized to match task complexity and latency budgets.

\subsection{Critical Gaps in Current Systems}
Despite significant advances in tool orchestration, our evaluation exposes two fundamental challenges that current frameworks cannot adequately address.

\textbf{Task Decomposition's Fragility:} Our failure analysis reveals that maintaining context across multi-step plans remains a critical weakness. The ``premature decomposition'' we observed leads to cascading errors, fundamentally limiting the ability to solve complex, long-horizon tasks that require sustained reasoning.

\textbf{Out-of-Scope Detection's Architectural Gap:} At Level 5, none of the evaluated orchestrators implements an explicit mechanism to detect when the available tool set cannot satisfy a query. In practice, correct rejection largely emerges from the backbone model's intrinsic reasoning rather than from purpose-built architectural checks. This dependency creates a safety gap: without architecture-level out-of-scope detection, rejection behavior is inconsistent across backbones and difficult to guarantee in deployment.

\subsection{Future Research Directions}
Based on these findings, we propose several promising directions for future research:
\begin{itemize}
    \item \textbf{Hierarchy-Aware Reasoning:} Develop reasoning and search strategies that go beyond using hierarchy as a simple filter, enabling models to explicitly leverage the semantic structure of tool servers to improve multi-level orchestration and inference.
    \item \textbf{Context-Propagating Decomposition:} Design new task decomposition methods that are explicitly engineered to maintain and pass global context between steps, preventing the context loss that plagues current models.
    \item \textbf{Adaptive and Hybrid Architectures:} Explore more sophisticated hybrid architectures that can dynamically choose between flat and hierarchical retrieval based on query complexity, or adaptively adjust retrieval strategies in real-time.
    \item \textbf{Developing Robust Rejection Mechanisms:} Create and integrate dedicated modules or prompting strategies specifically designed for out-of-scope detection, moving this critical safety feature from an emergent model property to a reliable architectural component.
\end{itemize}

%% file: sections/8_limitations.tex
\section{Limitations}
\label{sec:limitations}

Our framework presents three areas for future enhancement, each with clear rationale for the current design choices.

\textbf{Dataset Construction Methodology:} Our benchmark leverages proprietary LLMs (GPT-4.1~\cite{openai2024gpt4technicalreport}, Meta-Llama-3-8B-Instruct~\cite{grattafiori2024llama3herdmodels}) for data generation and human annotation for quality verification. While this approach requires computational resources (approximately \$500 USD for the complete benchmark), it ensures high-quality task generation that accurately reflects real-world multi-server orchestration challenges. We found that open-source alternatives, while cost-effective, produced less consistent results for our specific requirements. This investment in quality is justified by the benchmark's role as a standardized evaluation platform for the community.

\textbf{Simulation Fidelity vs. Orchestration Logic:} A key design choice in ETOM is the use of \emph{simulated execution} via pseudo-output schemas rather than live API calls. This ensures 100\% reproducibility and benchmark accessibility without requiring private platform keys. However, this means our simulations do not account for real-world stochasticity, such as random network latency, rate limits, or ``fuzzy'' unstructured content in tool responses. Consequently, ETOM is primarily a benchmark for orchestration logic and tool-use planning rather than an information extraction or noise-handling test. Future work could incorporate edge-case simulations, such as malformed tool outputs, to further test agent resilience.
\textbf{Evaluation Scope and Focus:} We prioritize end-to-end task completion metrics over granular reasoning trace analysis. While reasoning traces are valuable, focusing on end-to-end success provides a scalable and objective foundation for comparing diverse architectures. Our structured five-level curriculum and failure analysis (Section 5.5) already offer significant diagnostic insights, which can be further extended by future researchers using our released dataset.

\textbf{Dataset Diversity and Coverage:} Our benchmark draws from publicly available MCP servers on the glama.ai platform, representing current ecosystem trends rather than exhaustive coverage. This focused approach ensures that our evaluation reflects real-world deployment scenarios while maintaining manageable complexity. The English-based corpus captures the dominant language of current MCP implementations, and our server selection methodology prioritizes representative diversity over exhaustive coverage. Future extensions to multilingual scenarios and additional server sources represent natural evolution paths that build upon this foundation.

%% file: sections/9_ethical_considerations.tex
\section{Ethics Statement}
\label{sec:ethical_considerations}

Since our benchmark construction pipeline involves prompting LLMs for task generation, it is important to implement stringent measures to ensure the absence of offensive content in both the prompts and the generated responses. We first explicitly state in all generation prompts that the LLM should not generate any content that contains personal privacy violations, promotes violence, racial discrimination, hate speech, sexual content, or self-harm. Then, we manually inspect a random sample of 200 data entries from all five levels in ETOM for offensive content. Based on our observations, we have not detected any offensive content. Therefore, we believe that our benchmark is safe and will not yield any negative societal impact.

Due to data privacy and intellectual property concerns, our benchmark dataset will not be made public in its entirety. However, we provide detailed task generation pipelines and evaluation methodologies to ensure reproducibility. For language models, we access all open-source LMs via the Hugging Face Hub. The number of parameters is presented in our experimental results. All associated licenses permit user access for research purposes, and we have agreed to follow all terms of use.

We conduct human annotations for quality verification and task validation. Our annotators are graduate students and research assistants who are compensated at standard academic rates. The selection of annotators is based on their domain expertise in AI systems and tool orchestration, and we do not collect any personal information beyond what is necessary for the annotation process. All annotators agree to participate as their contribution to the research without additional compensation beyond their standard academic roles.

Our benchmark assumes access to diverse external tools and services, which may not be available to all users due to economic, geographical, or institutional constraints. This could potentially exacerbate digital inequalities. We acknowledge this limitation and encourage future work to consider accessibility and inclusivity in tool orchestration system design and deployment.

We used ChatGPT and Gemini to assist in proofreading and code documentation during the writing process.

\section*{Acknowledgments}

This work was primarily conducted during an internship at Delta Electronics, Inc. We gratefully acknowledge their support, which made this research possible. We would like to express our sincere gratitude to our mentors—Jimmy, TinTin, Alex, Leo, and Kun-Tai—for their invaluable guidance and insightful discussions. We also thank the anonymous reviewers for their constructive feedback and suggestions, which significantly helped improve the depth and clarity of this work. Finally, our gratitude extends to the broader MCP community and the glama.ai~\cite{glama_ai} platform, whose efforts in consolidating tool servers provided the essential foundation for our benchmark corpus.

%% file: sections/A1_algorithm_details.tex
\section{Algorithm for Equal Function Set Generation}
\label{app:algorithm_details}

This appendix provides the core algorithmic description for Equal Function Set Generation, which serves as the foundation for Level 2 and Level 4 ground truth annotation in ETOM. The algorithm (Algorithm~\ref{alg:lv2_rag}) implements a two-phase approach: bottom-up candidate generation and verification, followed by top-down query-guided RAG verification to establish functionally equivalent tool sets.

\begin{algorithm*}[ht]
\caption{Level 2 Query-Tool Round-Trip Consistency Generation}
\label{alg:lv2_rag}
\KwIn{Tool corpus $\mathcal{T}$, similarity threshold $\tau$, Level 2 queries $\mathcal{Q}$}
\KwOut{Verified Level 2 query-tool associations}

\BlankLine
\textbf{/* Phase 1: Bottom-up Candidate Generation and Verification */}
\BlankLine

$P \gets \varnothing$ \tcp{Set of unique candidate pairs}
$V \gets \varnothing$ \tcp{Set of verified equivalent pairs}

\BlankLine
\textbf{/* 1a. Generate candidate pairs from semantic similarity */}
\ForEach{tool $t \in \mathcal{T}$}{
    $C_t \gets$ top-10 semantically similar tools to $t$ with similarity $> \tau$\;
    \ForEach{$(t_i, t_j) \in C_t$ \textbf{where} $t_i \neq t_j$}{
        $(a, b) \gets (\min(t_i, t_j), \max(t_i, t_j))$\;
        $P \gets P \cup \{(a,b)\}$\;
    }
}

\BlankLine
\textbf{/* 1b. Verify functional equivalence of pairs via LLM */}
\ForEach{$(t_i, t_j) \in P$}{
    \If{\textsc{LlmEquiv}$(t_i, t_j) = \text{TRUE}$}{
        $V \gets V \cup \{(t_i, t_j)\}$\;
    }
}

\BlankLine
\textbf{/* 1c. Build Equal Function Sets using Union-Find */}
Let $U$ be a Union--Find data structure initialized with all tools in $\mathcal{T}$\;
\ForEach{$(t_i, t_j) \in V$}{
    \textsc{Union}$(U, t_i, t_j)$\;
}
Equal function sets $\gets$ root groups from $U$\;

\BlankLine
\textbf{/* Phase 2: Top-down Query-Guided RAG Verification */}
\ForEach{query $q \in \mathcal{Q}$}{
    $R \gets$ Retrieve top-10 relevant tools for $q$ via RAG\;
    $S \gets$ Select tools from $R$ that can fulfill $q$ via LLM\;
    Human-verify consistency of $S$ against the Equal Function Sets\;
}

\BlankLine
\Return{Verified Level 2 query-tool associations}
\end{algorithm*}

%% file: sections/A2_task_generation_pipelines.tex
\section{ETOM Task Generation Pipelines}
\label{app:task_generation_pipelines}

This appendix details the multi-stage, semi-automated pipelines for generating ETOM tasks across Levels 1–-5 (see Figure~\ref{fig:five_level_curriculum} for the overall curriculum design). The goal is to produce high-quality, verifiable queries that systematically evaluate distinct agent capabilities.

\subsection{Server Filtering Methodology}
\label{appendix:method}

Before task generation begins, we adopted a two-stage process for filtering MCP servers to ensure only high-quality, relevant servers are included in our corpus:

\begin{enumerate}
    \item \textbf{LLM-based Filtering}: We used an LLM(GPT-4.1~\cite{openai2024gpt4technicalreport}) to assess whether each MCP server provided genuine, indispensable external capabilities beyond native LLM functions. The full prompt text is provided in Appendix~\ref{app:server_filtering_prompt} for reproducibility.
    \item \textbf{Human Review}: Borderline cases were manually examined to ensure accuracy and consistency with the defined criteria.
\end{enumerate}

\subsection{MCP Server Data Format Example}
\label{app:mcp_data_format}

\begin{figure}[H]
    \centering
    \begin{tcolorbox}[
        enhanced,
        title={\bfseries MCP Server Data Format Example},
        left=6mm, right=6mm, top=4mm, bottom=4mm,
        fonttitle=\bfseries, colback=gray!5, colframe=gray!75
    ]
    \raggedright
    \begin{itemize}
        \item \textbf{Server Name:} \texttt{heroku-mcp-server}
        \item \textbf{Description:} Heroku Platform MCP Server for LLM interaction with Heroku resources
        \item \textbf{URL:} \url{https://glama.ai/mcp/servers/@heroku/heroku-mcp-server}
        \item \textbf{Summary:} MCP implementation for seamless LLM-Heroku Platform interaction
        \item \textbf{Categories:} \texttt{agent-orchestration}, \texttt{cloud-platform}
        
        \item \textbf{Tools:}
        \begin{itemize}
            \item \textbf{Tool Name:} \texttt{rename\_app}
            \item \textbf{Description:} Change Heroku app name or resolve naming conflicts
            \item \textbf{Input Schema:}
            \begin{itemize}
                \item \textbf{Properties:}
                \begin{itemize}
                    \item \texttt{app} (string): Current name of the Heroku app to rename
                    \item \texttt{newName} (string): New unique name for the app
                \end{itemize}
                \item \textbf{Required Fields:} \texttt{app}, \texttt{newName}
                \item \textbf{Type:} object
            \end{itemize}
        \end{itemize}
    \end{itemize}
    \end{tcolorbox}
    
    \caption{
        \textbf{MCP Server Data Format:} Standard JSON structure for MCP server data used in ETOM, containing server metadata, tool descriptions, and input schemas required for the task generation pipeline.
    }
    \label{fig:mcp_data_format}
\end{figure}
\vspace{-1.5em}

This format provides all necessary information for the tool triage and annotation process, including server metadata, tool descriptions, and input schemas required for the ETOM task generation pipeline.

\subsection{Tool Triage and Semantic Annotation}
Before query generation, all tools in the corpus undergo rigorous triage and annotation to filter for suitable candidates and collect metadata. We employ targeted LLM classifiers for:

\paragraph{1. Platform Identification} 
Identify the primary user-facing platform, brand, or product associated with each server using \texttt{GPT-4.1}~\cite{openai2024gpt4technicalreport}. 

\paragraph{2. Task Type Classification} 
Classify each tool as \texttt{final\_goal} or \texttt{middleware} using Meta-Llama-3-8B-Instruct~\cite{grattafiori2024llama3herdmodels}.

\paragraph{3. User Orientation Classification} 
Determine if a tool is \texttt{user\_facing} or \texttt{system\_facing} using \texttt{GPT-4.1}~\cite{openai2024gpt4technicalreport}. 

Only tools with a specific platform, classified as \texttt{final\_goal} and \texttt{user\_facing}, are eligible for query generation. Prompt details are available in Appendix~\ref{app:server_filtering_prompt}.

\subsection{Level 1 Pipeline: Foundational Tasks}
\label{subsec:l1_pipeline}

\paragraph{Stage 1: Generation} 
\seqsplit{Meta-Llama-3-8B-Instruct}~\cite{grattafiori2024llama3herdmodels} generates two direct, explicit user commands per tool, mentioning the platform and including concrete examples for tool parameters in case follow-up questions arise (e.g., ``Which document do you want to modify?''). The detailed prompt is provided in Appendix~\ref{app:prompts_l1_generation}.

\paragraph{Stage 2: Verification} 
A second LLM validates the generated queries for intent match, mandatory platform mention, and lack of ambiguity. Invalid queries are discarded. The verification prompt is given in Appendix~\ref{app:prompts_l1_verification}.

\subsection{Level 2 Pipeline: Context-Aware Tasks}
\label{subsec:l2_pipeline}

\sloppy
\paragraph{Stage 1: Platform Coupling Analysis} 
Classify tools as \texttt{tightly\_coupled} (platform-specific) or \texttt{generic\_concept} using \texttt{Meta-Llama-3-8B-Instruct}~\cite{grattafiori2024llama3herdmodels}. Classification prompt is in Appendix~\ref{app:prompts_l2_coupling}.
\fussy

\paragraph{Stage 2: Dynamic Rule Generation \& Query Generation} 
Set the \texttt{platform\_mention\_rule} as:
\begin{itemize}
    \item For \texttt{tightly\_coupled} tools: ``You MUST mention the platform \texttt{{platform\_name}} because the function is iconic to it.''
    \item For \texttt{generic\_concept} tools: ``You MUST NOT mention the platform \texttt{{platform\_name}}, as the function is a generic concept.''
\end{itemize}
Rule-based query generation prompt: see Appendix~\ref{app:prompts_l2_query}.

\paragraph{Stage 3: Verification} 
Generated L2 queries are verified by \texttt{Meta-Llama-3-8B-Instruct}~\cite{grattafiori2024llama3herdmodels} against the rules above. Verifier prompt: Appendix~\ref{app:prompts_l2_verification}.

\paragraph{Stage 4: Round-Trip Consistency Validation}
\label{subsec:stage4_validation}
This stage balances LLM scalability with human precision to establish a high-fidelity ground truth. The validation follows a two-stage logic to refine the Equal Function Sets (EFS) introduced in Section~\ref{subsec:equal_function_sets}:
\begin{enumerate}
\item \textbf{Bottom-Up Discovery:} An initial broad discovery phase where an LLM analyzes all tool descriptions across 491 servers to propose candidate sets of functionally similar tools. This stage yielded 145 initial candidate sets.
\item \textbf{Top-Down Query-Conditioned Validation:} For each candidate set, we retrieve the top-10 relevant tools via RAG using the generated Level 2 query. An LLM (\texttt{GPT-4.1}) identifies tools capable of solving the specific query.
\item \textbf{Platform-Aware Human Refinement:} Human validators (10 student-hours) manually examine borderline cases to distinguish between textual similarity and true platform interchangeability.

\textit{Example:} In one instance, the initial set grouped \texttt{Notion::list\_note} and \texttt{HackMD::list\_note} based on textual similarity. However, when tested against a HackMD-specific query (e.g., ``Find my documents in HackMD''), human validators rejected the Notion tool as irrelevant and instead consolidated \texttt{HackMD::list\_note} and \texttt{HackMD::list\_document} into a single, platform-consistent set.

\end{enumerate}
Through this rigorous validation, the initial 145 candidate sets were refined into the final 95 high-confidence Equal Function Sets that form the L2 ground truth. LLM selection prompt: Appendix~\ref{app:prompts_l2_roundtrip}.

\subsection{Level 3 Pipeline: Intra-Server Sequential Chaining}
\label{subsec:l3_pipeline}

\sloppy
\paragraph{Stage 1: Inferring Pseudo Output Schemas} 
Meta-Llama-3-8B-Instruct~\cite{grattafiori2024llama3herdmodels} infers plausible output JSON schemas for each tool based on its name, description, and inputs. The full prompt is in Appendix~\ref{app:prompts_l3_infer_schema}.
\fussy

\paragraph{Stage 2: Graph Construction \& Chain ID} 
Construct a directed graph $G=(V,E)$, where nodes are tools and edges represent valid input-output flows. Identify candidate chains as simple paths in the graph.

\sloppy
\paragraph{Stage 3: Two-Stage Query Generation} 
\begin{enumerate}
    \item \textbf{Logical Verification:} A verifier LLM checks whether the chain represents a logical, realistic workflow. Prompt in Appendix~\ref{app:prompts_l3_logical_verif}.
    \item \textbf{Query and CoT Generation:} Generator LLM produces high-level queries and Chain-of-Thought plans for verified chains. Prompt in Appendix~\ref{app:prompts_l3_query_cot}.
\end{enumerate}
\fussy

\subsection{Level 4 Pipeline: Cross-Server Compositional Chaining}
\label{subsec:l4_pipeline}

\begin{enumerate}
    \item \textbf{Server Sampling:} Group 491 servers into 20+ functional categories; randomly sample 2~4 categories and select one server from each category to ensure cross-server composition.
    
    \item \textbf{Feasibility Check:} Use \texttt{Qwen3-4B-Instruct}~\cite{yang2025qwen3technicalreport} to evaluate whether the sampled server combination can form a logical user workflow. The LLM determines feasibility using structured output with \seqsplit{\texttt{<is\_feasible>true/false\allowbreak</is\_feasible>}} tags and provides reasoning for its decision. Prompt in Appendix~\ref{app:prompts_l4_feasibility}.
    
    \item \textbf{Workflow Generation:} If feasible, use \texttt{GPT-4.1}~\cite{openai2024gpt4technicalreport} to generate a realistic user query and corresponding tool workflow with explicit dependencies. The output includes:
    \begin{itemize}
        \item Natural language query requiring cross-server coordination
        \item Tool sequence with dependency mapping (using array indices)
        \item Tool IDs in format \texttt{server\_name::tool\_name}
        \item Parallel execution support for independent tools
    \end{itemize}
    Prompt in Appendix~\ref{app:prompts_l4_workflow_generation}.
    
    \item \textbf{Quality Control:} Multi-dimensional evaluation using \texttt{GPT-4.1}~\cite{openai2024gpt4technicalreport} to assess query quality and ensure parameter completeness. We adapt the Task Quality Assessment Prompt methodology from \citet{wang2025mcpbenchbenchmarkingtoolusingllm} to ensure queries fully correspond to tools and contain all necessary parameter information, avoiding LLM follow-up questions. The evaluation assesses:
    \begin{itemize}
        \item \textbf{Parameter Completeness (1-10):} Whether the query contains all necessary parameters for tool execution
        \item \textbf{Naturalness (1-10):} Whether the query maintains conversational style vs. task-list format
    \end{itemize}
    Iterative improvement with up to 3 retries, accepting queries with parameter completeness $\geq 7.0$. Prompt in Appendix~\ref{app:prompts_l4_quality_control}.
    
    \item \textbf{Human Verification:} Each generated task is manually validated one-by-one for logical coherence, plausibility, and executability.
    
    \item \textbf{Equal Function Set Integration:} Following human verification, we map each ground truth tool in the generated tasks to corresponding Equal Function Sets (EFS) to ensure comprehensive ground truth coverage. For each tool position, we identify all functionally equivalent alternatives from the EFS and optionally verify their applicability in the specific query context using multi-model LLM voting (Qwen3-4B~\cite{yang2025qwen3technicalreport}, Llama-3-8B~\cite{grattafiori2024llama3herdmodels}, GPT-4.1~\cite{openai2024gpt4technicalreport}). This process ensures that the ground truth tool paths contain all feasible solutions rather than single tool instances. Tool verification prompt in Appendix~\ref{app:prompts_l4_efs_verification}.
\end{enumerate}

\subsection{Level 5 Pipeline: Robustness via Capability Gap Identification}
\label{sec:appendix_l5}

\paragraph{Phase 1: Capability Mapping} 
Tool description embeddings are clustered via HDBSCAN and labeled by LLMs to map available capabilities. Prompt in Appendix~\ref{app:prompts_l5_capability_mapping}.

\paragraph{Phase 2: Gap Analysis} 
An agentic debate framework identifies functional gaps:
\begin{enumerate}
    \item ``Proposer'' \texttt{Meta-Llama-3-8B-Instruct}~\cite{grattafiori2024llama3herdmodels} brainstorms universal user tasks. 
    Prompt in Appendix~\ref{app:prompts_l5_proposer}.
    \item Check solvability with existing capabilities.
    \item ``Red Team'' \texttt{GPT-4.1}~\cite{openai2024gpt4technicalreport} attempts to solve each task. 
    Prompt in Appendix~\ref{app:prompts_l5_redteam}.
    \item Confirm a capability gap if both fail.
\end{enumerate}

\paragraph{Phase 3: Persona-Based Query Generation} 
Generate queries from diverse user personas (e.g., ``Business Analyst,'' ``Student'') for each gap. Prompt in Appendix~\ref{app:prompts_l5_persona_query}.

\paragraph{Phase 4: Final Verification} 
Retrieve semantically similar tools and use a ``Judge'' agent to confirm no tool can solve the query. Only definitively out-of-scope queries are included. Prompt in Appendix~\ref{app:prompts_l5_final_verification}.


%% file: sections/B1_implementation_details.tex
\section{Orchestrator Implementation Details}
\label{app:impl_details}

This appendix provides implementation details for the orchestrator architectures evaluated in ETOM. For each methodology, we outline the concrete implementation choices, deviations from the original frameworks, and adaptations required to integrate with our benchmark’s MCP-based tool ecosystem.

\subsection{ReAct Orchestrator}
\paragraph{Overview} 
The ReAct framework~\cite{yao2023reactsynergizingreasoningacting} is designed around an iterative cycle of \textbf{Reasoning}, \textbf{Action}, and \textbf{Observation}. In its canonical form, an agent produces intermediate reasoning traces, executes actions (tool calls), and incorporates observations from the environment into subsequent reasoning steps. This iterative loop enables the agent to solve multi-step problems and recover from errors.

\paragraph{Adaptations for ETOM}  
In our benchmark, real tool execution is not performed. Instead, we modified the observation stage to align with the MCP tool ecosystem:
\begin{itemize}[topsep=0pt, itemsep=2pt]
    \item \textbf{Server/Tool Validation:} When the agent selects a server and tool for execution, the “observation” is defined as a lookup against the MCP server pool. If the chosen tool exists in the specified server, the execution is considered a success.
    \item \textbf{Simulated Execution Feedback:} Successful validation produces a structured record of the tool call (tool name, server name, input schema). This serves as the observation that is passed back into the next reasoning cycle.
    \item \textbf{Failure Handling:} If the tool cannot be matched in the MCP dataset, the step is marked as failed and the orchestrator continues the cycle with the updated context.
\end{itemize}
This design preserves the iterative “reason–act–observe” nature of ReAct while ensuring evaluation is feasible in a purely offline benchmark setting.

\paragraph{Multi-Attempt Variant (ReAct@\emph{N})}  
We extend the original design to allow multiple independent reasoning trajectories for robustness. In ReAct@\emph{N}, the orchestrator generates up to \(N\) distinct solution paths (via diversified orchestration prompts). The overall attempt is deemed successful if at least one trajectory leads to a valid tool orchestration. This extension provides a systematic way to evaluate the sensitivity of ReAct-style agents to orchestration variance.

\paragraph{Configuration}
The ReAct implementation in ETOM uses two role-based LLMs:
\begin{itemize}[topsep=0pt, itemsep=2pt]
    \item \textbf{Router LLM:} Handles query classification, server ranking, tool ranking, and orchestration prompts.
    \item \textbf{Conversational LLM:} Generates direct responses for queries determined not to require tool usage (L5 scenarios).
\end{itemize}
Key configurable parameters include the number of servers to shortlist (\texttt{server\_top\_k}), the number of tools per server (\texttt{tool\_top\_k}), and the maximum number of reasoning cycles per attempt. For ReAct@\emph{N}, we set \(N \in \{1, 3, 5\}\) in line with our experimental setup.

\paragraph{Error Handling and Level 5 Detection} 
To ensure consistent benchmarking, we implemented explicit error classification (e.g., \\ \texttt{server\_selection\_error}, \\ \texttt{tool\_selection\_error}, \texttt{orchestration\_error}). Additionally, we introduce a post-hoc check to distinguish between genuine Level~5 cases (queries not requiring tools) and process failures. This detection relies on analyzing the reasoning text and the absence of attempted tool calls.

\paragraph{Prompting Structure}
The ReAct orchestrator uses structured prompting to enforce consistent outputs across stages, including prompts for server selection, tool selection, orchestration, and error handling. 
For example, the server selection stage requires a JSON array of server names. 
The full set of ReAct prompts is provided in Appendix~\ref{app:prompts}, see in particular:
\begin{itemize}[topsep=0pt, itemsep=2pt]
    \item Server Selection Prompt (\ref{app:react_server_prompt})
    \item Tool Selection from Servers Prompt (\ref{app:react_tool_selection_prompt})
    \item Needle Selection Prompt (\ref{app:react_needle_prompt})
    \item Orchestration Prompt (\ref{app:react_planning_prompt})
\end{itemize}

\subsection{ToolShed Orchestrator}
\paragraph{Overview} 
The ToolShed framework~\cite{lumer2024toolshedscaletoolequippedagents} is designed to address the challenge of large-scale tool retrieval by combining enhanced tool representations with advanced Retrieval-Augmented Generation (RAG) techniques. ToolShed introduces the \textbf{ToolShed Knowledge Base (TKB)}, a vector database containing enriched tool documents, and an \textbf{Advanced RAG-Tool Fusion} pipeline that balances retrieval accuracy and efficiency. The pipeline consists of three major phases: pre-retrieval enrichment, intra-retrieval query transformation and retrieval, and post-retrieval reranking.

\paragraph{Adaptations for ETOM}
In our benchmark, we implement ToolShed as a high-performance baseline for flat tool retrieval, with several adaptations to fit the MCP ecosystem and our evaluation requirements:
\begin{itemize}[topsep=0pt, itemsep=2pt]
    \item \textbf{Pre-retrieval:} We adopt the original ToolShed pre-retrieval methodology during data preparation by enhancing MCP server and tool embeddings with metadata, hypothetical queries, and topical expansions. This ensures compatibility with the vector-based retrieval stage.
    \item \textbf{Tool-vs-Conversation Classification:} Unlike the original ToolShed pipeline, we prepend a classification stage to determine whether a query requires tool usage. Queries classified as \emph{conversational} are redirected to a standard LLM response pathway (Level~5 handling).
    \item \textbf{Intra-retrieval:} The intra-retrieval phase follows the ToolShed design, including:
    \begin{enumerate}[topsep=0pt, itemsep=2pt]
        \item \emph{Query Rewriting} for normalization and intent clarification.
        \item \emph{Query Expansion and Decomposition} into multiple semantically diverse variations.
        \item \emph{Multi-query Retrieval} from the MCP-enhanced tool pool using embedding similarity.
    \end{enumerate}
    \item \textbf{Post-retrieval:} Retrieved candidates are deduplicated and reranked using an LLM-based reranker conditioned on both the original query and expanded variations. This step mirrors ToolShed’s Advanced RAG-Tool Fusion.
    \item \textbf{Self-Reflection for Robustness:} We extend the original framework by incorporating a self-reflection step that allows the orchestrator to reject infeasible or out-of-scope requests, aligning with Level~5 evaluation in ETOM.
\end{itemize}

\paragraph{Configuration} 
The ToolShed orchestrator requires multiple role-specialized LLMs, including:
\begin{itemize}[topsep=0pt, itemsep=2pt]
    \item \textbf{Router:} Classifies queries and oversees final tool selection.
    \item \textbf{Query Rewriter:} Normalizes and clarifies input queries.
    \item \textbf{Query Expander:} Generates multiple diverse query variations.
    \item \textbf{Decomposer:} Produces structured sub-queries when appropriate.
    \item \textbf{Reranker:} Reorders candidate tools using cross-query context.
    \item \textbf{Conversational:} Provides direct responses for non-tool queries.
\end{itemize}
Key parameters include the number of query variations generated (\texttt{query\_expansion\_count}), the similarity threshold for embedding retrieval, and the number of top candidates reranked (\texttt{rerank\_top\_k}).

\paragraph{Error Handling and Level 5 Detection}
To support our evaluation protocol, the orchestrator is instrumented with explicit error logging and classification (e.g., decomposition errors, retrieval failures, reranking format errors). Out-of-scope or unfulfillable queries are rejected through the self-reflection mechanism, which outputs the standardized Level~5 rejection tuple (\texttt{\{server: "No", tool: "No"\}}).

\paragraph{Prompting Structure}
ToolShed’s implementation uses specialized prompts at each retrieval stage, including query rewriting, expansion, decomposition, and reranking. 
The full set of ToolShed prompts is provided in Appendix~\ref{app:prompts}, see:
\begin{itemize}[topsep=0pt, itemsep=2pt]
    \item Rewrite Prompt (\ref{app:toolshed_rewrite_prompt})
    \item Query Expansion Prompt (\ref{app:toolshed_expansion_prompt})
    \item Rerank Prompt (\ref{app:toolshed_rerank_prompt})
    \item Shared prompts: Classification (\ref{app:classification_prompt}), 
          Tool Selection (\ref{app:tool_selection_prompt}), 
          Query Decomposition (\ref{app:query_decomposition_prompt})
\end{itemize}

\subsection{MCP-Zero Orchestrator}
\paragraph{Overview} 
MCP-Zero~\cite{fei2025mcpzeroactivetooldiscovery} proposes a framework for \emph{active tool discovery}, enabling LLM agents to dynamically request tools on demand rather than relying on static schema injection. This design reduces context overhead, improves scalability to thousands of tools, and grants agents autonomy to specify capability gaps explicitly. The original methodology combines:
\begin{itemize}[topsep=0pt, itemsep=2pt]
    \item \textbf{Active Tool Requests:} LLMs generate structured requests specifying both the server domain and tool operation.
    \item \textbf{Hierarchical Semantic Routing:} A two-stage retrieval process that first matches servers by embeddings and then ranks tools within the top servers.
    \item \textbf{Iterative Invocation:} Theoretically supports multi-turn refinement of tool requests when initial retrieval is insufficient.
\end{itemize}

\paragraph{Adaptations for ETOM} 
Our implementation of MCP-Zero preserves its hierarchical retrieval philosophy but introduces several practical adaptations for benchmark evaluation:
\begin{itemize}[topsep=0pt, itemsep=2pt]
    \item \textbf{Tool-vs-Conversation Classification:} As with ToolShed, we prepend a classification stage to determine whether a query requires tools or can be directly answered conversationally.
    \item \textbf{Multi-Turn Decomposition:} While the original MCP-Zero mentions \emph{Iterative Active Invocation}, concrete implementation details were not available in the paper or reference repository. To operationalize this, we integrate the query decomposition mechanism from ToolShed. The decomposer LLM generates sub-queries, which are then independently processed through the MCP-Zero retrieval pipeline.
    \item \textbf{Enhanced Embeddings:} Instead of raw embeddings, we leverage the pre-retrieval enhanced embeddings from the ToolShed-prepared MCP server/tool pool, ensuring more robust semantic matching.
    \item \textbf{Self-Reflection for Robustness:} A reflection stage is included to allow graceful rejection of out-of-scope queries, aligning with Level~5 evaluation.
\end{itemize}

\paragraph{Configuration} 
The MCP-Zero orchestrator relies on three essential LLM roles, with an additional decomposer to enable multi-turn capability:
\begin{itemize}[topsep=0pt, itemsep=2pt]
    \item \textbf{Router:} Classifies queries and selects the final tool.
    \item \textbf{Retriever:} Generates structured tool requests in the \texttt{<tool\_assistant>} format.
    \item \textbf{Conversational:} Provides direct answers for queries not requiring tool usage.
    \item \textbf{Decomposer:} Expands queries into sub-queries to support multi-turn retrieval.
\end{itemize}
Key hyperparameters include the similarity threshold for embeddings, the number of servers shortlisted (\texttt{server\_top\_k}), and the number of tools ranked within each server (\texttt{tool\_top\_k}).

\paragraph{Error Handling and Level 5 Detection} 
Similar to ToolShed, explicit error categories (classification errors, retrieval failures, selection errors) are logged to ensure interpretability of benchmark results. Self-reflection enables the orchestrator to reject infeasible tasks, producing the standardized rejection tuple \texttt{\{server: "No", tool: "No"\}}.

\paragraph{Prompting Structure} 
MCP-Zero uses structured prompts to elicit tool requests and tool selections. 
The key prompt is the Tool Transformation Prompt (\ref{app:mcpzero_tooltrans_prompt}), which ensures the retriever LLM emits structured \texttt{<tool\_assistant>} requests. 
In addition, MCP-Zero uses the shared classification, decomposition, and tool selection prompts listed in Appendix~\ref{app:prompts}.

\subsection{Hybrid Orchestrator}
\paragraph{Overview} 
The Hybrid orchestrator represents an experimental design introduced in ETOM to explore the potential synergy between the \textbf{active tool request formulation} of MCP-Zero and the \textbf{advanced retrieval pipeline} of ToolShed. The motivation is to test whether combining MCP-Zero’s structure-aware tool request transformation with ToolShed’s powerful multi-query retrieval and reranking mechanisms yields improvements in robustness and accuracy.

\paragraph{Architecture} 
The Hybrid pipeline largely follows the three-phase structure of ToolShed, with one critical substitution:
\begin{itemize}[topsep=0pt, itemsep=2pt]
    \item \textbf{Stage 1 (Pre-Processing):} Instead of performing query rewriting as in ToolShed, the Hybrid agent invokes MCP-Zero’s \emph{Tool Transformation Prompt} to convert the raw user query into a structured tool request containing explicit \texttt{<server>} and \texttt{<tool>} descriptors.
    \item \textbf{Stage 2 (Intra-Retrieval):} The structured request is then passed into ToolShed’s query decomposition and expansion modules. Multiple query variations are generated, embeddings are computed, and tools are retrieved across the MCP server/tool pool.
    \item \textbf{Stage 3 (Post-Retrieval):} Candidate tools are deduplicated and reranked using ToolShed’s LLM-based reranking stage, with the top-$k$ tools selected for final evaluation.
\end{itemize}

\paragraph{Adaptations for ETOM} 
\begin{itemize}[topsep=0pt, itemsep=2pt]
    \item Added a \textbf{tool-vs-conversation classification} step (same as other orchestrators) to bypass retrieval when direct conversational answers suffice.
    \item Integrated the \textbf{Tool Transformation Prompt} (from MCP-Zero) into the pre-retrieval stage, replacing ToolShed’s rewrite prompt. This ensures that queries are reformulated in a structured, server/tool-aware manner before expansion and retrieval.
    \item Leveraged the \textbf{enhanced embeddings} of the ToolShed-prepared MCP dataset for retrieval.
    \item Included the same \textbf{self-reflection mechanism} as other orchestrators, allowing the agent to reject out-of-scope queries and align with Level~5 evaluation.
\end{itemize}

\paragraph{Configuration} 
The Hybrid orchestrator uses an extended set of LLM roles:
\begin{itemize}[topsep=0pt, itemsep=2pt]
    \item \textbf{Router:} Handles classification and final tool selection.
    \item \textbf{Retriever:} Performs tool request transformation (adopted from MCP-Zero).
    \item \textbf{Query Expander:} Generates multiple variations of the transformed request.
    \item \textbf{Reranker:} Reorders candidate tools after multi-query retrieval.
    \item \textbf{Conversational:} Responds to non-tool queries.
    \item \textbf{Decomposer:} Enables multi-turn sub-query processing for complex requests.
\end{itemize}
Key parameters mirror those of ToolShed, with \texttt{query\_expansion\_count} and \texttt{rerank\_top\_k} controlling the breadth of retrieval and reranking depth.

\paragraph{Prompting Structure}
The Hybrid orchestrator combines MCP-Zero’s structured tool transformation with ToolShed’s query expansion and reranking. 
It therefore uses the Tool Transformation Prompt (\ref{app:mcpzero_tooltrans_prompt}), along with the ToolShed expansion (\ref{app:hybrid_expansion_prompt}) and rerank prompts (\ref{app:hybrid_rerank_prompt}), plus the shared classification, decomposition, and tool selection prompts in Appendix~\ref{app:prompts}.

\paragraph{Remarks} 
This design allows us to empirically test whether explicit, structured tool requests (MCP-Zero) can improve downstream retrieval quality when combined with ToolShed’s RAG-Tool Fusion. As an experimental variant, it is not intended as a new SOTA baseline, but rather as a probe into potential complementarities between structure-aware and retrieval-enhanced paradigms.

%% file: sections/B2_evaluation_framework.tex
\section{Evaluation Framework}
\label{app:evaluation_framework}

This appendix provides a comprehensive evaluation framework for ETOM, including detailed protocols, metric definitions, scoring mechanisms, and validation procedures for each level.

\subsection{Level-Specific Evaluation Protocols}
\label{app:level_protocols}

\noindent\textbf{Level 1 (L1): Foundational Tool Identification}
\label{app:l1_protocol}

\paragraph{Objective} 
Establish a baseline for mapping a direct, unambiguous user directive to the correct tool within a specified server.

\paragraph{Evaluation Process} 
\begin{enumerate}[topsep=0pt, itemsep=2pt]
    \item The agent receives the full server specification, including tool descriptions and input schemas
    \item The agent must identify the single correct tool for the given query
    \item Correctness is strictly based on exact identity with the ground-truth tool tuple
    \item No partial credit is awarded for similar or functionally equivalent tools
\end{enumerate}

\paragraph{Example}
\begin{itemize}[topsep=0pt, itemsep=2pt]
    \item \textbf{Query:} "Run a Semgrep scan on vulnerable\_code.js with config.yaml"
    \item \textbf{Expected:} \texttt{semgrep\_scan} from \texttt{Semgrep MCP Server}
    \item \textbf{Correct:} Exact match with tool name and server
\end{itemize}

\noindent\textbf{Level 2 (L2): Disambiguation Among Functionally Equivalent Tools}
\label{app:l2_protocol}

\paragraph{Objective} 
Assess reasoning under functional redundancy, where multiple tools can fulfill the same request.

\paragraph{Evaluation Process} 
\begin{enumerate}[topsep=0pt, itemsep=2pt]
    \item The agent receives the query and candidate tools from the specified server
    \item Multiple tools may be functionally equivalent for the given task
    \item Predictions are scored against the ground-truth set of valid tools
    \item Emphasis is on correctness of selection over preference
\end{enumerate}

\paragraph{Example}
\begin{itemize}[topsep=0pt, itemsep=2pt]
    \item \textbf{Query:} "Convert this image to JPEG format"
    \item \textbf{Valid Tools:} \{\texttt{convert\_to\_jpeg},\\ \texttt{image\_format\_converter}\}
    \item \textbf{Correct:} Either tool from the valid set
\end{itemize}

\noindent\textbf{Level 3 (L3): Intra-Server Task Orchestration}
\label{app:l3_protocol}

\paragraph{Objective} 
Measure the agent's capacity to decompose a high-level goal into a coherent, executable plan using tools within a single server.

\paragraph{Evaluation Process} 
\begin{enumerate}[topsep=0pt, itemsep=2pt]
    \item Agent must construct a DAG representing the execution plan
    \item All tools must belong to the same server
    \item Dependencies must be correctly ordered
    \item Both exact sequence matching and tool selection accuracy are evaluated
\end{enumerate}

\paragraph{Example}
\begin{itemize}[topsep=0pt, itemsep=2pt]
    \item \textbf{Query:} "Analyze code for vulnerabilities and create a report"
    \item \textbf{Expected Plan:} \texttt{semgrep\_scan} $\rightarrow$ \seqsplit{\texttt{generate\_report}} (both from Semgrep server)
    \item \textbf{Evaluation:} Exact sequence and tools within single server
\end{itemize}

\noindent\textbf{Level 4 (L4): Cross-Server Task Orchestration}
\label{app:l4_protocol}

\paragraph{Objective} 
Measure the agent's capacity to decompose a high-level goal into a coherent, executable plan using tools across multiple servers.

\paragraph{Evaluation Process} 
\begin{enumerate}[topsep=0pt, itemsep=2pt]
    \item Agent must construct a DAG with tools from different servers
    \item Correct server identification is required
    \item Cross-server dependencies must be properly ordered
    \item Both exact sequence matching and tool selection accuracy are evaluated
\end{enumerate}

\paragraph{Example}
\begin{itemize}[topsep=0pt, itemsep=2pt]
    \item \textbf{Query:} "Scan code for vulnerabilities, then email the results to the team"
    \item \textbf{Expected Plan:} \texttt{semgrep\_scan} (Semgrep server) $\rightarrow$ \texttt{send\_email} (Email server)
    \item \textbf{Evaluation:} Exact sequence, tools, and server identification
\end{itemize}

\noindent\textbf{Level 5 (L5): Rejection of Unfulfillable Requests}
\label{app:l5_protocol}

\paragraph{Objective} 
Evaluate the agent's ability to correctly reject requests impossible to fulfill within the MCP ecosystem.

\paragraph{Evaluation Process} 
\begin{enumerate}[topsep=0pt, itemsep=2pt]
    \item The ground truth for L5 is an empty execution plan
    \item Correct rejection requires the exact tuple \seqsplit{\texttt{\{server: "no", tool: "no"\}}}
    \item Any other output is considered a failure
    \item Ensures the metric captures judgment of feasibility rather than operational errors
\end{enumerate}

\paragraph{Example}
\begin{itemize}[topsep=0pt, itemsep=2pt]
    \item \textbf{Query:} "Turn on the lights in my living room"
    \item \textbf{Expected:} \seqsplit{\texttt{\{server: "no", tool: "no"\}}}
    \item \textbf{Correct:} Proper rejection of physical world interaction
\end{itemize}

%% file: sections/C1_additional_statistics.tex
\section{Additional Statistics}
\label{app:additional_statistics}

This appendix provides detailed statistical visualizations and distribution charts that support the main analysis in Section~\ref{sec:benchmark_statistics}.

\subsection{Category Distribution}
\label{app:category_dist}

The distribution of server categories in the ETOM corpus reflects the diverse ecosystem of available MCP servers. This visualization shows how tools are distributed across different functional domains.

\begin{figure}[ht]
    \centering
    \includegraphics[width=0.8\linewidth]{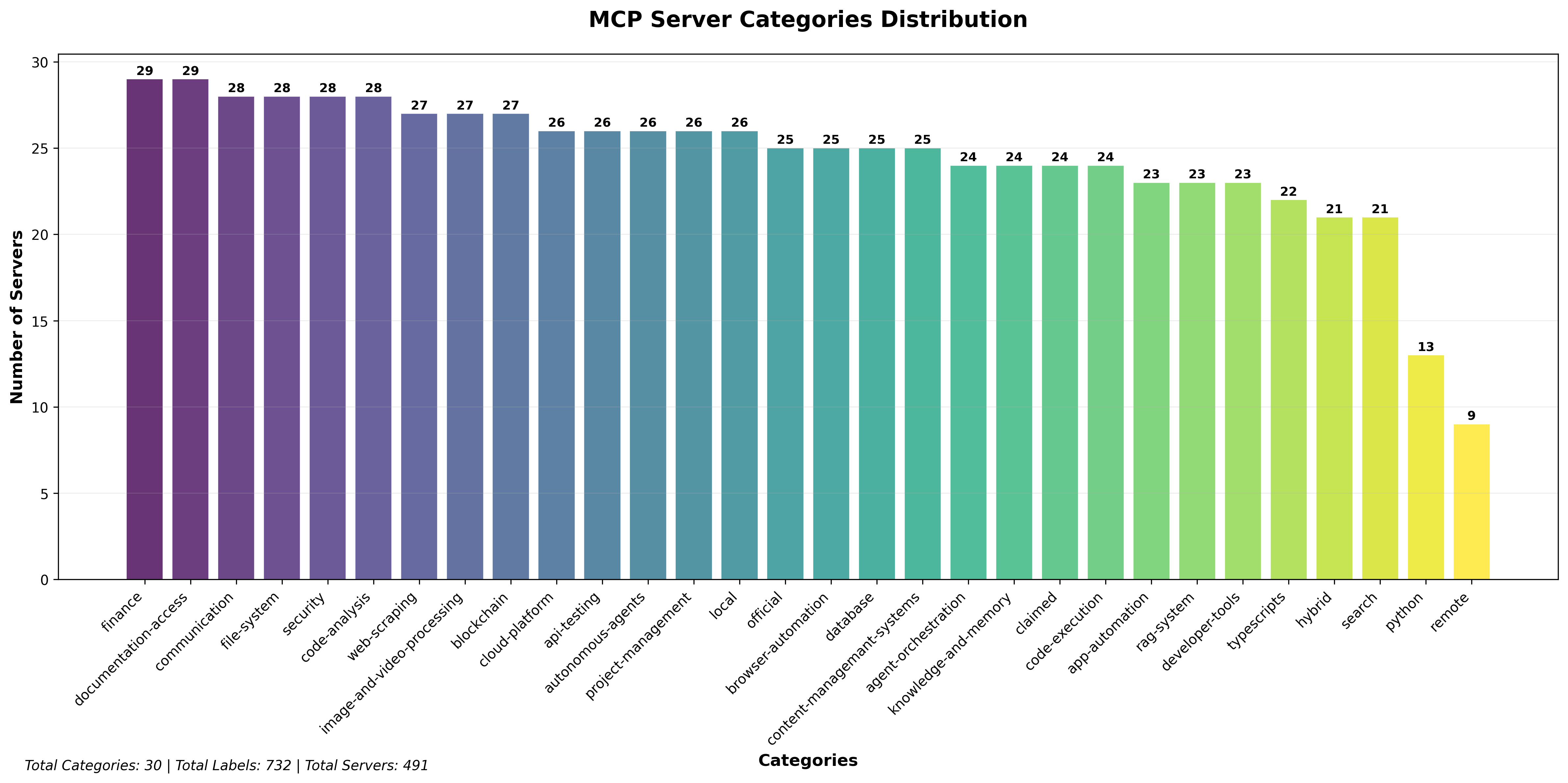}
    \caption{Distribution of server categories in the ETOM corpus. The 30 categories represent diverse functional domains, from development tools to productivity applications.}
    \label{fig:category_dist_appendix}
\vspace{-1.5em}
\end{figure}

\subsection{Retrieval Complexity Comparison} 
\label{app:complexity_comparison}

Table~\ref{tab:complexity_comparison_appendix} provides a detailed comparison of retrieval complexity between ETOM's foundational levels (L1, L2) and other prominent single-step benchmarks, based on lexical (ROUGE-L) and semantic (METEOR) similarity between queries and tool descriptions.

\begin{table}[h] 
    \centering
    \small 
    \begin{tabular}{lccc}
        \toprule
        \textbf{Benchmark} & \textbf{Type} & \textbf{ROUGE-L} & \textbf{METEOR} \\
        \midrule
        \multicolumn{4}{l}{\textit{Existing Benchmarks}} \\
        API-Bank & Intent-based & 0.116 & 0.095 \\
        ToolE & Intent-based & 0.106 & 0.102 \\
        ToolRet & Intent-based & 0.148 & 0.178 \\
        \midrule
        \multicolumn{4}{l}{\textit{ETOM}} \\
        \textbf{L1} & \textbf{Command-based} & \textbf{0.337} & \textbf{0.337} \\
        \textbf{L2} & \textbf{Intent-based} & \textbf{0.152} & \textbf{0.076} \\
        \bottomrule
    \end{tabular}
    \caption{Retrieval complexity comparison. ETOM spans both high-overlap (L1) and low-overlap (L2) tasks.}
    \label{tab:complexity_comparison_appendix} 
\end{table}

\subsection{Level 5 Capability Gap Analysis}
\label{app:level5_dist}

The systematic identification of capability gaps ensures comprehensive robustness testing across different domains. This distribution shows how the 91 out-of-scope queries are distributed across 31 distinct capability gap categories.

\begin{figure}[ht]
    \centering
    \includegraphics[width=0.9\columnwidth]{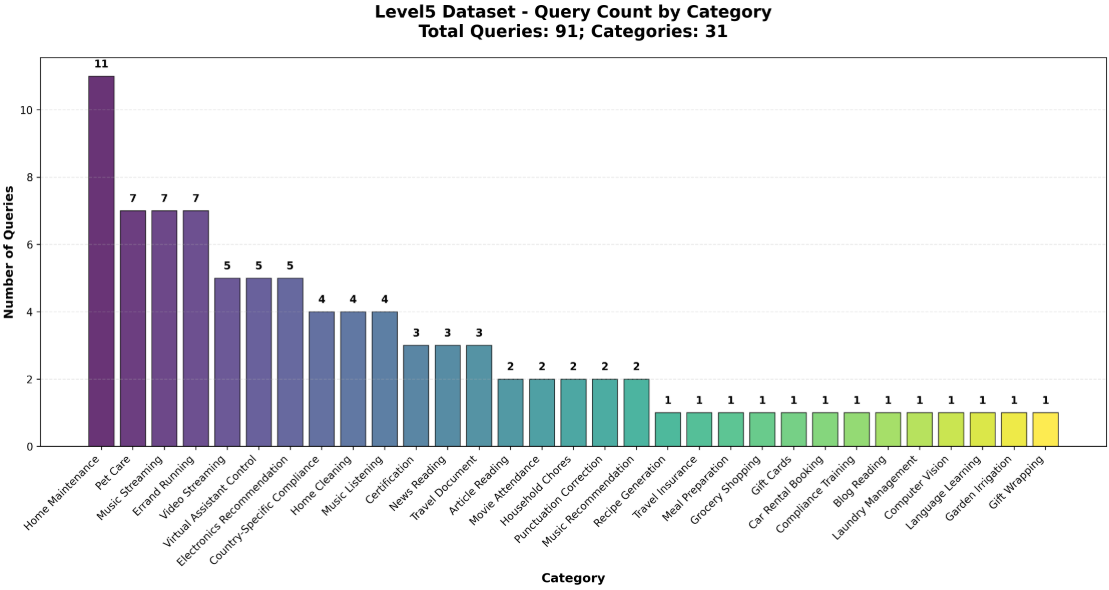} 
    \caption{Distribution of 31 capability gap categories for Level 5 out-of-scope queries. Each category represents a well-defined domain outside the digital MCP ecosystem boundaries.}
    \label{fig:level5_distribution_appendix}
\vspace{-1.5em}
\end{figure}

\noindent\textbf{Capability Gap Categories}
The 31 capability gap categories represent well-defined domains outside the digital MCP ecosystem boundaries, systematically organized to ensure comprehensive robustness testing:

\paragraph{Physical World Interaction (28 queries, 31\%)}
\begin{itemize}[topsep=0pt, itemsep=2pt]
    \item \textbf{Smart Home Control:} Controlling lights, temperature, security systems
    \item \textbf{Physical Device Control:} Controlling printers, scanners, cameras, IoT devices
\end{itemize}

\paragraph{Real-Time Sensory Processing (22 queries, 24\%)}
\begin{itemize}[topsep=0pt, itemsep=2pt]
    \item \textbf{Audio Processing:} Real-time speech recognition, audio analysis, voice commands
    \item \textbf{Visual Processing:} Real-time video analysis, object recognition, facial detection
\end{itemize}

\noindent\textbf{Agent Judgment Requirements}
For Level 5 queries, agents must demonstrate robust capability boundary detection by:

\begin{enumerate}[topsep=0pt, itemsep=2pt]
    \item \textbf{Comprehensive Tool Review:} Examining all available tools across the entire corpus to determine if any tool can address the query
    \item \textbf{Capability Boundary Recognition:} Understanding the fundamental limitations of the digital MCP ecosystem
    \item \textbf{Confident Rejection:} Only rejecting queries after thorough verification that no tool exists to solve them
    \item \textbf{Clear Communication:} Responding with \texttt{\{server: "no", tool: "no"\}} when definitively out-of-scope
\end{enumerate}

This design ensures that agents cannot simply guess or make assumptions about capability limitations, but must systematically verify tool availability before concluding that a query is unfulfillable.

%% file: sections/D1_detailed_eval_result.tex
\begin{table*}[!t]
\centering
\footnotesize
\begin{tabular}{cccccccc}
\toprule
\multicolumn{3}{c}{\textbf{Configuration}} & \textbf{L1} & \textbf{L2} & \textbf{L3} & \textbf{L4} & \textbf{L5} \\
\cmidrule(lr){1-3} \cmidrule(lr){4-4} \cmidrule(lr){5-5} \cmidrule(lr){6-6} \cmidrule(lr){7-7} \cmidrule(lr){8-8}
\textbf{k} & \textbf{QE} & \textbf{RR} & \textbf{EM} & \textbf{EM} & \textbf{EM} & \textbf{F1} & \textbf{EM} \\
\midrule
5  & 3 & 10 & 57.03 & 42.12 & 46.17 & 50.77 & 14.56 \\
5  & 0 & 10 & 57.03 & 41.60 & 50.76 & 47.11 & 11.65 \\
1  & 0 & 10 & 45.78 & 36.56 & 36.08 & 37.91 & 4.85 \\
5  & 3 & 1  & 54.09 & 42.50 & 44.95 & 46.78 & 10.67 \\
10 & 3 & 10 & 56.64 & 40.69 & 47.40 & 50.95 & 13.59 \\
15 & 3 & 10 & 55.88 & 43.15 & 48.31 & 49.55 & 14.56 \\
20 & 3 & 10 & 57.80 & 35.91 & 50.15 & \textbf{52.11} & 10.67 \\
25 & 3 & 10 & 56.01 & 33.72 & 47.70 & 45.54 & 12.62 \\
\bottomrule
\end{tabular}
\caption{Complete ToolShed ablation study results on Meta-Llama-3-8B-Instruct~\cite{grattafiori2024llama3herdmodels}. \textbf{Configuration}: k = \texttt{tool\_top\_k} (retrieval breadth), QE = \texttt{query\_expansion\_count}, RR = \texttt{rerank\_top\_k}. \textbf{Metrics}: EM = Exact Match (\%), F1 = F1 Score (\%) for L4 only. Bold values indicate optimal performance within each metric.}
\label{tab:toolshed_complete_ablation}
\end{table*}

\section{Detailed Evaluation Results}
\label{app:detailed_evaluation_results}

This appendix provides comprehensive experimental results for the ToolShed ablation study, including all tested parameter configurations and their corresponding performance metrics across evaluation levels L1-L5.

\section{Detailed Ablation Study of ToolShed’s Architecture}
\label{app:ablation}

To thoroughly investigate the impact of its core components, we performed an extensive ablation study on the ToolShed architecture using the Meta-\allowbreak Llama-\allowbreak 3-\allowbreak 8B-\allowbreak Instruct model~\cite{grattafiori2024llama3herdmodels}. We systematically varied three key parameters:
\begin{itemize}
    \item \textbf{Retrieval Breadth (\texttt{tool\_top\_k})}: The number of initial candidate tools retrieved.
    \item \textbf{Query Expansion (\texttt{query\_expansion\_count})}: Whether to use LLM-based query expansion to broaden the search.
    \item \textbf{Reranking (\texttt{rerank\_top\_k})}: The number of candidates considered by the reranker.
\end{itemize}

\begin{figure}[h!]
\centering
\includegraphics[width=0.9\linewidth]{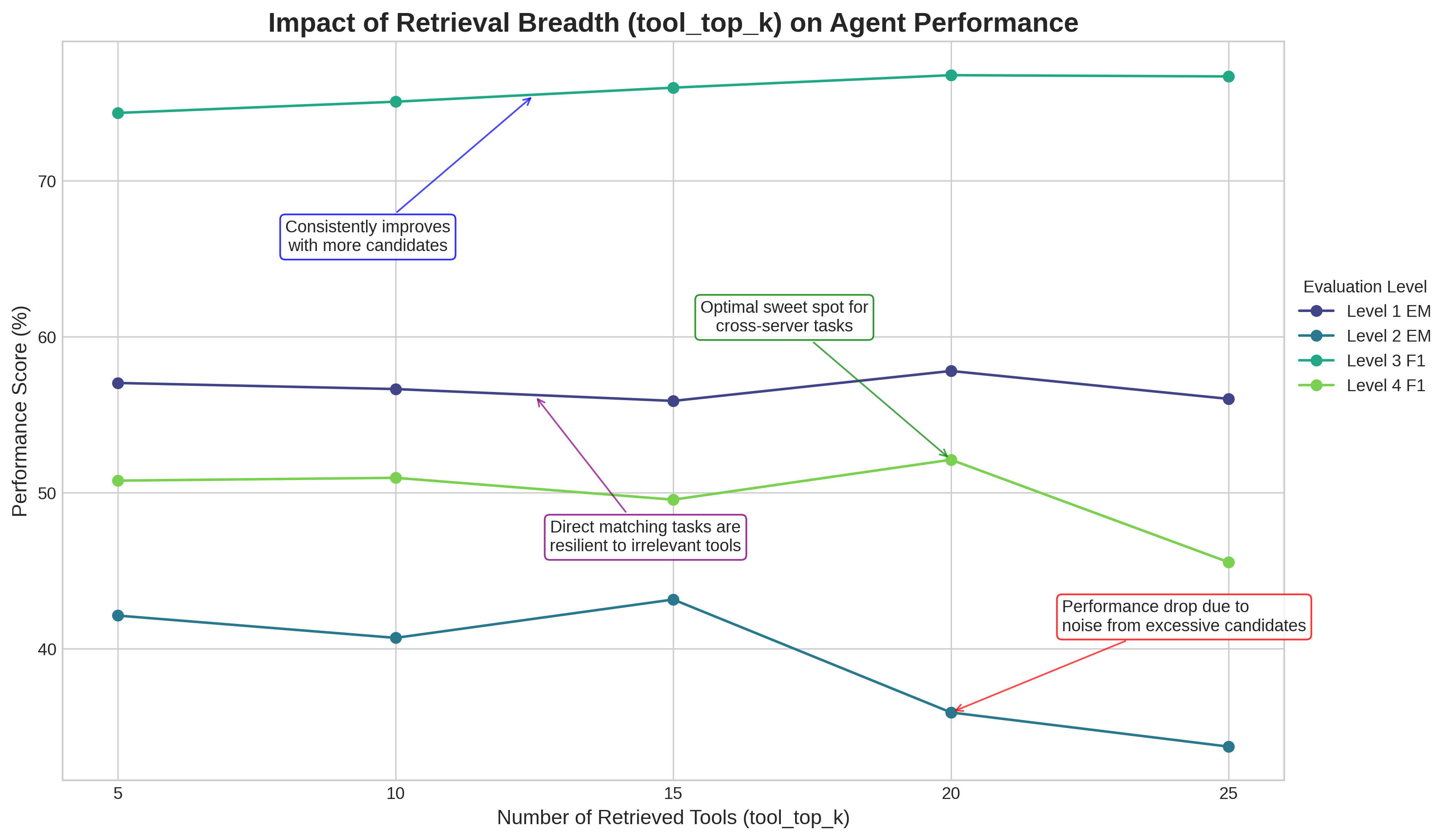}
\caption{Impact of retrieval breadth (tool\_top\_k) on ToolShed performance across complexity levels. L1/L2 show EM scores, L3/L4 show F1 scores. Annotations highlight key behavioral patterns: L1 resilience to irrelevant tools, L2 degradation from excessive candidates, L3 consistent improvement, and L4 optimal performance at k=20.}
\label{fig:toolshed_retrieval_breadth}
\vspace{-1em}
\end{figure}

Table~\ref{tab:toolshed_ablation} provides a summary of key configurations, while Table~\ref{tab:toolshed_complete_ablation} presents the complete numerical results. Figure~\ref{fig:toolshed_retrieval_breadth} visualizes the impact of retrieval breadth.

\begin{table}[h!]
\centering
\small
\begin{tabular}{lccc}
\toprule
\textbf{Setting} & \textbf{Retrieval} & \textbf{QE} & \textbf{L2/L3/L4} \\
\midrule
Full & 5 & Yes & 42.12/74.34/50.77 \\
No QE & 5 & No & 41.60/\textbf{76.98}/47.11 \\
Narrow & 1 & No & 36.56/62.50/37.91 \\
Wide & 20 & Yes & 35.91/76.76/\textbf{52.11} \\
No Rerank & 5 & Yes & 42.50/71.91/46.78 \\
\bottomrule
\end{tabular}
\caption{ToolShed ablation study on Llama. Retrieval = tool\_top\_k, QE = query expansion.}
\label{tab:toolshed_ablation}
\vspace{-1em}
\end{table}

Table~\ref{tab:toolshed_complete_ablation} presents the complete ablation study results for ToolShed on Meta-Llama-3-8B-Instruct~\cite{grattafiori2024llama3herdmodels}, systematically varying retrieval breadth (\texttt{tool\_top\_k}), query expansion (\seqsplit{\texttt{query\_expansion\_count}}), and reranking parameters (\texttt{rerank\_top\_k}).

\subsection{Key Observations and Analysis}

The complete results, summarized below, reveal several important patterns that inform the optimal configuration of a tool retrieval system.

\paragraph{1. Retrieval Breadth Effects are Task-Dependent}

Systematic variation of \texttt{tool\_top\_k} from 1 to 25 (visualized in Figure~\ref{fig:toolshed_retrieval_breadth}) shows a clear trade-off:
\begin{itemize}
    \item \textbf{L1 (Direct Matching)}: Performance is highly stable and peaks early (k=5), demonstrating strong resilience to irrelevant candidates in simple lookup tasks.
    \item \textbf{L2 (Context-Aware)}: Performance peaks at a medium breadth (k=\allowbreak 5,\allowbreak 43.15\%\allowbreak{} EM), after which it degrades due to noise from excessive, distracting tool candidates.
    \item \textbf{L3 (Intra-\allowbreak Server)}: Shows consistent improvement with increased breadth, as a wider view helps in complex single-server planning.
    \item \textbf{L4 (Cross-\allowbreak Server)}: Benefits the most from a wide retrieval scope, reaching optimal performance at k=\allowbreak 20\allowbreak{} (52.11\%\allowbreak{} F1). This confirms that complex, multi-domain orchestration requires access to a diverse toolset.
\end{itemize}

\paragraph{2. Query Expansion has a Nuanced Impact}
Direct comparison shows that Query Expansion (QE) is not universally beneficial:
\begin{itemize}
    \item It provides a notable benefit for the most complex textbf{L4 tasks} (50.77\% vs 47.11\% F1), where reformulating a query can help bridge different tool domains.
    \item It has minimal or even slightly negative impact on simpler textdf{L2 and L3 tasks}, where the original query is often sufficient and expansion can introduce noise.
\end{itemize}

\paragraph{3. Reranking is Critical for Quality}
Reducing the reranker's scope demonstrates its crucial role:
\begin{itemize}
    \item Performance degrades substantially across all levels when reranking is limited.
    \item This highlights that a powerful reranker is essential for filtering the initial, noisy set of retrieved candidates to find the most relevant tool.
\end{itemize}

These comprehensive results support the summary findings presented in Section~\ref{sec:ablation} and provide detailed evidence for our architectural choices.

\subsection{Comprehensive Performance and Efficiency Analysis}

See the main text Table~\ref{tab:appendix_full_results} for the complete experimental results across all architectures, foundation models, and evaluation levels, including both performance metrics and normalized latency measurements. This comprehensive view reveals several critical insights into the performance-efficiency trade-offs inherent in different tool orchestration approaches.

\paragraph{Architecture-Level Performance Patterns}
The results demonstrate clear architectural advantages across different complexity levels. ToolShed consistently outperforms MCP-Zero in raw performance metrics, achieving superior scores on complex orchestration tasks (L3: 84.34\% vs 53.97\% F1 peak, L4: 55.06\% vs 34.20\% F1 peak). However, this performance advantage comes at a substantial computational cost, with ToolShed configurations showing 3-20× higher normalized latency compared to their MCP-Zero counterparts.

\paragraph{Foundation Model Sensitivity}
The comprehensive data reveals significant foundation model dependencies that extend beyond simple capability differences. For ToolShed architectures, GPT-4.1~\cite{openai2024gpt4technicalreport} achieves the highest performance on L4 cross-server orchestration (55.06\% F1) and L5 rejection tasks (81.31\% EM), while Meta-Llama-3-8B-Instruct~\cite{grattafiori2024llama3herdmodels} excels specifically on L3 intra-server coordination (84.34\% F1). Conversely, MCP-Zero shows more consistent performance patterns across models, with Qwen3-4B-Instruct-2507~\cite{yang2025qwen3technicalreport} demonstrating exceptional efficiency (0.22-6.50× normalized latency) while maintaining competitive accuracy.

\paragraph{Efficiency-Performance Trade-off Analysis}
The normalized latency measurements reveal a clear architectural trade-off: MCP-Zero's hierarchical approach achieves 2-10× better efficiency across most configurations, while ToolShed's dense retrieval and reranking pipeline delivers superior absolute performance at the cost of computational overhead. Notably, some ToolShed configurations (e.g., Microsoft Phi-4~\cite{abdin2024phi4technicalreport}) show both poor performance and high latency, indicating that the architecture's benefits are highly dependent on foundation model compatibility.

These detailed results provide the empirical foundation for the efficiency analysis presented in Section~\ref{sec:experiments} and demonstrate the complexity of balancing performance and computational costs in tool orchestration systems.

%% file: sections/D2_representative_examples.tex
\section{Representative Examples}
\label{app:representative_examples}

This appendix provides comprehensive representative examples for each level of ETOM, demonstrating the complexity and characteristics of tasks at each level.

\subsection{Level 1 Task Examples}
\label{app:l1_examples}

\begin{figure}[H]
    \centering
    \begin{tcolorbox}[
        enhanced,
        title={\bfseries Example: Level 1 (Direct Tool Retrieval)},
        left=6mm, right=6mm, top=4mm, bottom=4mm,
        fonttitle=\bfseries, colback=gray!5, colframe=gray!75
    ]
    \raggedright
    \begin{itemize}
        \item \textbf{User Query:} On Semgrep, run \texttt{semgrep\_scan} on the file \texttt{"vulnerable\_code.js"} with the configuration file \texttt{"semgrep\_config.yaml"} to detect vulnerabilities and return findings in JSON format.
        
        \item \textbf{Expected Tool:}
        \begin{itemize}
            \item \textbf{Server:} \texttt{Semgrep MCP Server}
            \item \textbf{Tool Name:} \texttt{semgrep\_scan}
            \item \textbf{Description:} Run static code analysis on provided files using Semgrep to detect vulnerabilities and return findings in JSON format for detailed inspection and remediation.
        \end{itemize}
    \end{itemize}
    \end{tcolorbox}
    
    \caption{
        \textbf{Level 1 Example:} Direct tool retrieval where the user query explicitly mentions the tool name and server. The task requires the agent to identify the exact tool from the specified server that matches the user's request.
    }
    \label{fig:l1_example}
\end{figure}
\vspace{-1.5em}

\begin{figure}[H]
    \centering
    \begin{tcolorbox}[
        enhanced,
        title={\bfseries Example: Level 1 (File Operations)},
        left=6mm, right=6mm, top=4mm, bottom=4mm,
        fonttitle=\bfseries, colback=gray!5, colframe=gray!75
    ]
    \raggedright
    \begin{itemize}
        \item \textbf{User Query:} Use the Filesystem MCP Server to create a new directory called \texttt{"project\_docs"} in the current working directory.
        
        \item \textbf{Expected Tool:}
        \begin{itemize}
            \item \textbf{Server:} \texttt{Filesystem MCP Server}
            \item \textbf{Tool Name:} \texttt{create\_directory}
            \item \textbf{Description:} Create a new directory at the specified path with the given name.
        \end{itemize}
    \end{itemize}
    \end{tcolorbox}
    
    \caption{
        \textbf{Level 1 Example:} Simple file system operation where the user clearly specifies the action and target. The agent must identify the correct tool for directory creation.
    }
    \label{fig:l1_example2}
\end{figure}

\subsection{Level 2 Task Examples}
\label{app:l2_examples}

\begin{figure}[H]
    \centering
    \begin{tcolorbox}[
        enhanced,
        title={\bfseries Example: Level 2 (Functionally Equivalent Tools)},
        left=6mm, right=6mm, top=4mm, bottom=4mm,
        fonttitle=\bfseries, colback=gray!5, colframe=gray!75
    ]
    \raggedright
    \begin{itemize}
        \item \textbf{User Query:} Convert this image to JPEG format with 90\% quality.
        
        \item \textbf{Valid Tools:}
        \begin{itemize}
            \item \textbf{Tool 1:} \texttt{convert\_to\_jpeg} - Convert images to JPEG format with specified quality
            \item \textbf{Tool 2:} \texttt{image\_format\_converter} - Convert images between different formats including JPEG
        \end{itemize}
        
        \item \textbf{Expected Response:} Either tool is acceptable as both can perform the requested conversion.
    \end{itemize}
    \end{tcolorbox}
    
    \caption{
        \textbf{Level 2 Example:} Ambiguous query where multiple tools can fulfill the same request. The agent must recognize that either tool is a valid choice for the conversion task.
    }
    \label{fig:l2_example}
\end{figure}
\vspace{-1.5em}

\begin{figure}[H]
    \centering
    \begin{tcolorbox}[
        enhanced,
        title={\bfseries Example: Level 2 (Multiple Search Tools)},
        left=6mm, right=6mm, top=4mm, bottom=4mm,
        fonttitle=\bfseries, colback=gray!5, colframe=gray!75
    ]
    \raggedright
    \begin{itemize}
        \item \textbf{User Query:} Search for information about "machine learning algorithms" in the documentation.
        
        \item \textbf{Valid Tools:}
        \begin{itemize}
            \item \textbf{Tool 1:} \texttt{search\_docs} - Search through documentation files
            \item \textbf{Tool 2:} \texttt{find\_content} - Find specific content within documents
            \item \textbf{Tool 3:} \texttt{query\_knowledge\_base} - Query the knowledge base for information
        \end{itemize}
        
        \item \textbf{Expected Response:} Any of the three tools can be used to search for the requested information.
    \end{itemize}
    \end{tcolorbox}
    
    \caption{
        \textbf{Level 2 Example:} Multiple functionally equivalent tools for searching documentation. The agent must understand that any of these tools can fulfill the search request.
    }
    \label{fig:l2_example2}
\end{figure}

\subsection{Level 3 Task Examples}
\label{app:l3_examples}

\begin{figure}[H]
    \centering
    \begin{tcolorbox}[
        enhanced,
        title={\bfseries Example: Level 3 (Single-Server Multi-Step Orchestration)},
        left=6mm, right=6mm, top=4mm, bottom=4mm,
        fonttitle=\bfseries, colback=gray!5, colframe=gray!75
    ]
    \raggedright
    \begin{itemize}
        \item \textbf{User Query:} Analyze the code in \texttt{src/main.py} for security vulnerabilities, generate a detailed report, and send it to the development team.
        
        \item \textbf{Expected Execution Plan:}
        \begin{enumerate}
            \item \textbf{Step 1:} \texttt{analyze\_code} - Scan the Python file for security issues
            \item \textbf{Step 2:} \texttt{generate\_report} - Create a detailed vulnerability report
            \item \textbf{Step 3:} \texttt{send\_notification} - Send the report to the development team
        \end{enumerate}
        
        \item \textbf{Dependencies:} Step 2 depends on Step 1, Step 3 depends on Step 2.
    \end{itemize}
    \end{tcolorbox}
    
    \caption{
        \textbf{Level 3 Example:} Multi-step orchestration within a single server (Security Analysis Server). The agent must decompose the high-level goal into a sequence of dependent tool calls.
    }
    \label{fig:l3_example}
\end{figure}
\vspace{-1.5em}

\begin{figure}[H]
    \centering
    \begin{tcolorbox}[
        enhanced,
        title={\bfseries Example: Level 3 (Database Operations)},
        left=6mm, right=6mm, top=4mm, bottom=4mm,
        fonttitle=\bfseries, colback=gray!5, colframe=gray!75
    ]
    \raggedright
    \begin{itemize}
        \item \textbf{User Query:} Create a new user account with the email "user@example.com", set up their profile, and send them a welcome email.
        
        \item \textbf{Expected Execution Plan:}
        \begin{enumerate}
            \item \textbf{Step 1:} \texttt{create\_user} - Create a new user account
            \item \textbf{Step 2:} \texttt{setup\_profile} - Configure user profile settings
            \item \textbf{Step 3:} \texttt{send\_welcome\_email} - Send welcome email to new user
        \end{enumerate}
        
        \item \textbf{Dependencies:} Step 2 depends on Step 1, Step 3 depends on Step 1.
    \end{itemize}
    \end{tcolorbox}
    
    \caption{
        \textbf{Level 3 Example:} User management workflow within a single server. The agent must plan the sequence of operations to complete the user onboarding process.
    }
    \label{fig:l3_example2}
\end{figure}

\subsection{Level 4 Task Examples}
\label{app:l4_examples}

\begin{figure}[H]
    \centering
    \begin{tcolorbox}[
        enhanced,
        title={\bfseries Example: Level 4 (Multi-Server Orchestration)},
        left=6mm, right=6mm, top=4mm, bottom=4mm,
        fonttitle=\bfseries, colback=gray!5, colframe=gray!75
    ]
    \raggedright
    \begin{itemize}
        \item \textbf{User Query:} Deploy the latest version of my web application to production, run security scans, and notify the team about the deployment status.
        
        \item \textbf{Expected Execution Plan:}
        \begin{enumerate}
            \item \textbf{Step 1:} \texttt{deploy\_app} (Deployment Server) - Deploy application to production
            \item \textbf{Step 2:} \texttt{run\_security\_scan} (Security Server) - Perform security assessment
            \item \textbf{Step 3:} \texttt{send\_notification} (Notification Server) - Notify team of deployment status
        \end{enumerate}
        
        \item \textbf{Dependencies:} Step 2 depends on Step 1, Step 3 depends on Steps 1 and 2.
    \end{itemize}
    \end{tcolorbox}
    
    \caption{
        \textbf{Level 4 Example:} Complex multi-server orchestration involving deployment, security, and notification services. The agent must coordinate tools across multiple servers.
    }
    \label{fig:l4_example}
\end{figure}
\vspace{-1.5em}

\begin{figure}[H]
    \centering
    \begin{tcolorbox}[
        enhanced,
        title={\bfseries Example: Level 4 (Data Pipeline)},
        left=6mm, right=6mm, top=4mm, bottom=4mm,
        fonttitle=\bfseries, colback=gray!5, colframe=gray!75
    ]
    \raggedright
    \begin{itemize}
        \item \textbf{User Query:} Extract data from the customer database, process it for analytics, and create visualizations for the quarterly report.
        
        \item \textbf{Expected Execution Plan:}
        \begin{enumerate}
            \item \textbf{Step 1:} \texttt{extract\_customer\_data} (Database Server) - Extract customer data
            \item \textbf{Step 2:} \texttt{process\_analytics} (Analytics Server) - Process data for analytics
            \item \textbf{Step 3:} \texttt{create\_visualizations} (Visualization Server) - Generate charts and graphs
            \item \textbf{Step 4:} \texttt{compile\_report} (Report Server) - Compile quarterly report
        \end{enumerate}
        
        \item \textbf{Dependencies:} Sequential dependencies across all steps.
    \end{itemize}
    \end{tcolorbox}
    
    \caption{
        \textbf{Level 4 Example:} Data processing pipeline involving multiple specialized servers. The agent must orchestrate a complex workflow across different services.
    }
    \label{fig:l4_example2}
\end{figure}

\subsection{Level 5 Task Examples}
\label{app:l5_examples}

\begin{figure}[H]
    \centering
    \begin{tcolorbox}[
        enhanced,
        title={\bfseries Example: Level 5 (Physical World Interaction)},
        left=6mm, right=6mm, top=4mm, bottom=4mm,
        fonttitle=\bfseries, colback=gray!5, colframe=gray!75
    ]
    \raggedright
    \begin{itemize}
        \item \textbf{User Query:} Turn on the lights in my living room and adjust the temperature to 22$^\circ$C.
        
        \item \textbf{Expected Response:} \seqsplit{\texttt{\{server: "no", tool: "no"\}}}
        
        \item \textbf{Reasoning:} This request requires physical world interaction (controlling lights and temperature) which is outside the scope of the digital MCP ecosystem. The agent should correctly reject this request.
    \end{itemize}
    \end{tcolorbox}
    
    \caption{
        \textbf{Level 5 Example:} Physical world interaction request that should be rejected. The agent must recognize that this type of request cannot be fulfilled within the digital tool ecosystem.
    }
    \label{fig:l5_example}
\end{figure}
\vspace{-1.5em}

\begin{figure}[H]
    \centering
    \begin{tcolorbox}[
        enhanced,
        title={\bfseries Example: Level 5 (Real-Time Sensory Processing)},
        left=6mm, right=6mm, top=4mm, bottom=4mm,
        fonttitle=\bfseries, colback=gray!5, colframe=gray!75
    ]
    \raggedright
    \begin{itemize}
        \item \textbf{User Query:} Listen to the conversation in the next room and summarize what they're discussing.
        
        \item \textbf{Expected Response:} \seqsplit{\texttt{\{server: "no", tool: "no"\}}}
        
        \item \textbf{Reasoning:} This request requires real-time audio processing and eavesdropping, which involves privacy concerns and real-time sensory capabilities not available in the digital MCP ecosystem.
    \end{itemize}
    \end{tcolorbox}
    
    \caption{
        \textbf{Level 5 Example:} Real-time sensory processing request that should be rejected. The agent must recognize privacy and capability limitations.
    }
    \label{fig:l5_example2}
\end{figure}
\vspace{-1.5em}

\subsection{Example Analysis}
\label{app:example_analysis}

\noindent\textbf{Complexity Progression}
The examples demonstrate a clear progression in complexity:

\begin{itemize}[topsep=0pt, itemsep=2pt]
    \item \textbf{Level 1:} Direct tool identification with explicit tool names
    \item \textbf{Level 2:} Ambiguous queries requiring reasoning about tool equivalence
    \item \textbf{Level 3:} Multi-step orchestration within a single server
    \item \textbf{Level 4:} Complex orchestration across multiple servers
    \item \textbf{Level 5:} Rejection of requests outside the digital ecosystem
\end{itemize}

\noindent\textbf{Key Characteristics}
Each level exhibits distinct characteristics:

\begin{itemize}[topsep=0pt, itemsep=2pt]
    \item \textbf{Level 1:} High precision requirements, explicit tool specification
    \item \textbf{Level 2:} Reasoning about functional equivalence, ambiguity handling
    \item \textbf{Level 3:} Task decomposition, dependency management
    \item \textbf{Level 4:} Cross-server coordination, complex workflow orchestration
    \item \textbf{Level 5:} Boundary recognition, appropriate rejection
\end{itemize}

\noindent\textbf{Evaluation Challenges}
The examples highlight key evaluation challenges:

\begin{itemize}[topsep=0pt, itemsep=2pt]
    \item \textbf{Level 1:} Exact matching requirements
    \item \textbf{Level 2:} Set-based evaluation, multiple valid answers
    \item \textbf{Level 3:} Dependency ordering, execution plan validation
    \item \textbf{Level 4:} Multi-server coordination, complex dependency graphs
    \item \textbf{Level 5:} Rejection accuracy, boundary understanding
\end{itemize}

%% file: sections/E1_prompt_templates.tex
\section{Prompt Templates}
\label{app:prompt_templates}

This appendix provides comprehensive prompt templates used throughout ETOM, including task generation prompts, orchestrator prompts, and evaluation prompts.

\subsection{Task Generation Prompts}
\label{app:task_generation_prompts}

\noindent\textbf{Server Filtering Prompt}
\label{app:server_filtering_prompt}

\begin{lstlisting}[
    basicstyle=\ttfamily\tiny,
    breaklines=true
]
You are a highly precise AI Benchmark Quality Analyst. Your mission is to evaluate a summarized MCP Server and classify it based on a strict set of rules. Your judgment must be consistent and based ONLY on the definitions provided.

## CRITICAL DEFINITION: What is "Native LLM Capability"?

For this task, a capability is **"native" if and only if** a Large Language Model can perform it using **solely its internal, pre-trained knowledge and reasoning abilities, without ANY access to external tools, APIs, or real-time data.**
Think of a "pure" LLM in a sandbox with no internet connection.

- **Native Examples:** Answering "What is the capital of France?", calculating 2+2, summarizing a provided text.
- **NON-Native Examples:** Accessing today's news (requires web search), creating a GitHub issue (requires API interaction), checking a stock price (requires real-time data).

Your entire evaluation MUST adhere to this strict definition.

## MCP Server Summary to Analyze
{mcp_server_json_content}

## REASONING & EVALUATION STEPS

1. **Analyze Core Functionality:** Based on the server's description and tools, what is its primary function?
2. **Apply the Critical Definition:** Can a pure LLM perform this function using only its internal knowledge?
3. **Make Your Classification:** Based on your analysis, classify this server.

## CLASSIFICATION RULES

**CLASSIFICATION: NATIVE**
- The server's primary function can be performed by a pure LLM using only internal knowledge
- Examples: Text processing, mathematical calculations, language translation, summarization

**CLASSIFICATION: NON-NATIVE**
- The server requires external tools, APIs, real-time data, or system access
- Examples: Web scraping, file operations, API calls, system commands, database access

## OUTPUT FORMAT

Provide your classification in exactly this format:

CLASSIFICATION: [NATIVE/NON-NATIVE]

REASONING: [Your detailed reasoning in 2-3 sentences explaining why you chose this classification]
\end{lstlisting}

\noindent\textbf{User Facing Classifier Prompt}
\label{app:User_Facing_Classifier_Prompt}

\begin{lstlisting}[
    basicstyle=\ttfamily\tiny,
    breaklines=true
]
# ROLE & GOAL
You are an AI Product Manager. Your task is to classify a given tool based on its intended user. You need to determine if it's a **"User-Facing Task"** or a **"System-Facing Task"**.

# DEFINITIONS
- **User-Facing Task**: An action that a typical end-user (like a writer, designer, project manager, or even a developer using a platform) would directly command an AI assistant to perform to achieve a personal or business goal. These tasks operate on user-understandable concepts like documents, repositories, images, emails, or playlists.
- **System-Facing Task**: An action related to system administration, infrastructure management, backend debugging, or managing abstract, non-visible resources. These tasks are typically performed by system administrators or developers maintaining a service, not using it. They operate on concepts like caches, database indexes, memory entries, or container pods.

# INSTRUCTIONS
1.  Read the tool's name and description carefully.
2.  Based on the definitions, decide if it's a "User-Facing Task" or a "System-Facing Task".
3.  Your output MUST be a single line containing the classification **wrapped in `<classification></classification>` tags**. The value must be either `user_facing` or `system_facing`.

# EXAMPLE 1
## Tool Name: "create_spreadsheet"
## Tool Description: "Creates a new spreadsheet in the user's cloud drive."
## Your Output:
<classification>user_facing</classification>

# EXAMPLE 2
## Tool Name: "clear_redis_cache"
## Tool Description: "Purges all keys from the specified Redis cache instance to free up memory."
## Your Output:
<classification>system_facing</classification>

# EXAMPLE 3
## Tool Name: "delete_memory_entry"
## Tool Description: "Deletes a specific memory entry for a user from the agent's long-term memory service."
## Your Output:
<classification>system_facing</classification>

---

# YOUR TASK
## Tool Name: "{tool_name}"
## Tool Description: "{tool_description}"
## Your Output:
\end{lstlisting}

\noindent\textbf{Platform ID Prompt}
\label{app:platform_id_prompt}

\begin{lstlisting}[
    basicstyle=\ttfamily\tiny,
    breaklines=true
]
PLATFORM_ID_PROMPT = """
# ROLE & GOAL
You are a highly intelligent text analysis engine. Your task is to identify the primary **user-facing Platform, Brand, or Product Keyword** from the provided text.

# INSTRUCTIONS
1.  Read the "Text to Analyze" carefully.
2.  Identify the main, specific, proper noun that represents the core service or brand being offered.
3.  This keyword should be a unique, user-recognizable identifier, like "GitHub", "Notion", "UseGrant", "Stripe", or "PaddleOCR".
4.  **Crucially, do NOT extract generic terms OR technology protocols.** This includes common words like "API", "Server", "Platform", "Service", and especially technical standards like **"MCP"** or **"Model Context Protocol"**.
5.  If you cannot find a specific, unique platform keyword after ignoring the terms above, the keyword is "N/A".
6.  Your output MUST be a single line containing the keyword **wrapped in `<keyword></keyword>` tags**.

# EXAMPLES
### Example 1 (Ignoring "MCP" and "API")
Text to Analyze: "This is a Model Context Protocol (MCP) server for interacting with the UseGrant API. It provides a set of tools for managing providers and clients through the UseGrant platform."
Your Output:
<keyword>UseGrant</keyword>

### Example 2 (Specific Technology Brand)
Text to Analyze: "A server for performing optical character recognition (OCR) using the powerful PaddleOCR engine."
Your Output:
<keyword>PaddleOCR</keyword>

### Example 3 (No Specific Brand Found)
Text to Analyze: "An MCP server that provides comprehensive architectural expertise through specialized agents, resources, and tools."
Your Output:
<keyword>N/A</keyword>

---

# YOUR TASK
**Text to Analyze**: "{text_to_analyze}"
**Your Output**:
\end{listing}

\noindent\textbf{Task Type Classification Prompt}
\label{app:task_type_prompt}

\begin{lstlisting}[
    basicstyle=\ttfamily\tiny,
    breaklines=true
]
# ROLE & GOAL
You are a pragmatic AI assistant designer. Your task is to analyze a given tool and classify its primary user intent. You need to determine if this tool represents a **"Final Goal Task"** or a **"Middleware Task"**.

# DEFINITIONS
- **Final Goal Task**: A task that a user would directly ask an AI assistant to perform as a complete, standalone goal. These tasks provide direct value to the user. (Examples: "search_for_videos", "download_a_file", "send_an_email", "translate_text").
- **Middleware Task**: A task that is usually an intermediate or prerequisite step required to achieve a larger goal. Users rarely, if ever, ask for this task directly. (Examples: "login", "authenticate", "get_api_key", "list_available_regions", "check_status").

# INSTRUCTIONS
1.  Read the tool's name and description.
2.  Based on the definitions, decide if it's a "Final Goal Task" or a "Middleware Task".
3.  Your output MUST be a single line containing the classification **wrapped in `<task_type></task_type>` tags**. The value must be either `final_goal` or `middleware`.

# EXAMPLE 1
## Tool Name: "search_youtube_videos"
## Tool Description: "Searches for videos on YouTube based on a query."
## Your Output:
<task_type>final_goal</task_type>

# EXAMPLE 2
## Tool Name: "youtube_login"
## Tool Description: "Authenticates the user and obtains an access token for the YouTube API."
## Your Output:
<task_type>middleware</task_type>

---

# YOUR TASK
## Tool Name: "{tool_name}"
## Tool Description: "{tool_description}"
## Your Output:
\end{lstlisting}

\noindent\textbf{Level 1 Generation Prompt}
\label{app:prompts_l1_generation}

\begin{lstlisting}[
    basicstyle=\ttfamily\tiny,
    breaklines=true
]
# ROLE & GOAL
You are an expert user of a specific software tool. Your task is to write 2 user queries in English. These queries should be direct commands to an AI assistant, demonstrating how a real user would request to use the specified tool on its target platform. The queries must be specific, self-contained, and unambiguous.

# CONTEXT FOR YOUR TASK
You are generating queries for the following specific tool:

- **Platform Name**: "{platform_name}" 
  (The user-facing brand or service, e.g., "GitHub", "Heroku", "Pyodide")
- **Server Name**: "{server_name}" 
  (The name of the software package providing the tool, e.g., "github-mcp-server")
- **Tool Name**: "{tool_name}"
  (The specific function to be executed, e.g., "create_repository")
- **Tool Description**: "{tool_description}"
  (What the tool does in plain English)

# INSTRUCTIONS
1.  **Direct Command Tone**: Your queries should be direct commands, not questions.
2.  **Mention Platform**: Each query MUST explicitly mention the **Platform Name** ("{platform_name}"). This is crucial for context.
3.  **Reflect Tool's Function**: The command's intent MUST be a direct application of the **Tool Description** and its **Tool Inputs**.
4.  **AVOID AMBIGUOUS REFERENCES (VERY IMPORTANT)**:
    - Do NOT use vague pointers like "this code", "that file", "the script".
    - For tools that execute code, embed a short, realistic code snippet directly in the query.
    - For tools that use files, specify a plausible filename.
5.  **Be Specific**: Incorporate realistic example values for the parameters listed in **Tool Inputs**. This makes the query more concrete and useful for training.
6.  **Format**: You MUST wrap each generated query in `<query></query>` tags.

# EXAMPLE
## Context:
- **Platform Name**: "GitHub"
- **Server Name**: "github-mcp-server"
- **Tool Name**: "create_repository"
- **Tool Description**: "Creates a new repository on GitHub."

## Your Output:
<query>On GitHub, create a new private repository named 'my-secret-project' with the description 'This is for the new API'.</query>
<query>Use GitHub to create a public repository called 'awesome-list'.</query>

---

# YOUR TASK
Now, using the context provided at the top, generate 2 user queries following all the rules.

# YOUR OUTPUT
\end{lstlisting}

\noindent\textbf{Level 1 Verification Prompt}
\label{app:prompts_l1_verification}

\begin{lstlisting}[
    basicstyle=\ttfamily\tiny,
    breaklines=true
]
# ROLE & GOAL
You are a meticulous AI System Analyst. Your task is to perform a strict validation of a user query against a specific tool. The query must be a perfect example of a command a user would give for this tool.

# CONTEXT
- **Platform Name**: "{platform_name}"
- **Tool Name**: "{tool_name}"
- **Tool Description**: "{tool_description}"

# VALIDATION CRITERIA
You must check the query against ALL of the following rules. If ANY rule is violated, the result is `false`.

1.  **Intent Match**: Does the user's primary goal in the query directly and logically map to the tool's function described in the **Tool Description**?
2.  **Platform Mention**: Does the query EXPLICITLY mention the required **Platform Name** ("{platform_name}")?
3.  **Self-Contained & Unambiguous**: Is the query understandable on its own without needing prior conversation? It must NOT use vague references like "this code", "that file" unless it also provides a concrete example (e.g., embedding code, providing a filename).
4.  **No Meta-Language**: Does the query sound like a real user? It must NOT refer to the tool itself (e.g., "use this tool", "run the function"). The command must be direct.

# YOUR TASK
Analyze the query below based on all the criteria.

## Query to Verify: "{query}"

# YOUR RESPONSE
Structure your output as follows:
- First, provide the boolean judgement (`true` or `false`) **wrapped in `<is_match></is_match>` tags**.
- Second, on a new line, provide a one-sentence explanation for your decision, specifically mentioning which rule was passed or failed.

# EXAMPLE 1: Perfect Match (All Rules Pass)
## Context:
- Platform Name: "GitHub"
- Tool Name: "create_repository"
- Tool Description: "Creates a new repository on GitHub."
## Query to Verify: "On GitHub, please create a new repository for me named 'my-next-project'."
## Your Output:
<is_match>true</is_match>
The query matches the tool's intent, mentions the platform 'GitHub', is self-contained, and uses natural user language.

# EXAMPLE 2: Failed (No Platform Mention)
## Context:
- Platform Name: "GitHub"
- Tool Name: "create_repository"
- Tool Description: "Creates a new repository on GitHub."
## Query to Verify: "Create a new repository for me named 'my-next-project'."
## Your Output:
<is_match>false</is_match>
The query fails validation because it does not explicitly mention the required platform 'GitHub'.

# EXAMPLE 3: Failed (Ambiguous Reference)
## Context:
- Platform Name: "Pyodide"
- Tool Name: "execute-python"
- Tool Description: "Executes a string of Python code."
## Query to Verify: "Run this Python code using Pyodide."
## Your Output:
<is_match>false</is_match>
The query fails validation because "this Python code" is an ambiguous reference, violating the self-contained rule.

# EXAMPLE 4: Failed (Meta-Language)
## Context:
- Platform Name: "LocalFS"
- Tool Name: "delete_file"
- Tool Description: "Deletes a file from the filesystem."
## Query to Verify: "Use the LocalFS tool to delete the file 'temp.log'."
## Your Output:
<is_match>false</is_match>
The query fails validation because "Use the LocalFS tool" is meta-language, not natural user language.
\end{lstlisting}

\noindent\textbf{Level 2 Platform Coupling Classification Prompt}
\label{app:prompts_l2_coupling}

\begin{lstlisting}[
    basicstyle=\ttfamily\tiny,
    breaklines=true
]
# ROLE & GOAL
You are a Product Analyst specializing in software tools and user behavior. Your task is to determine if a tool's core function is tightly coupled with its specific platform, or if it represents a generic concept that exists across many platforms.

# DEFINITIONS
- **Tightly Coupled**: The tool's main concept or terminology is unique to its platform and doesn't make sense outside of it. A user would HAVE to mention the platform name to be understood.
  - *Examples*: "Managing Heroku Dynos", "Creating a GitHub Pull Request", "Resolving a Jira Transition". The concepts "Dyno", "Pull Request", and "Jira Transition" are iconic to their platforms.
- **Generic Concept**: The tool's function is a common action that many different platforms offer under similar names. A user could describe this task without mentioning a specific brand.
  - *Examples*: "Creating a file", "Sending an email", "Uploading an image", "Renaming a project".

# INSTRUCTIONS
1.  Analyze the tool's name, description, and the platform it belongs to.
2.  Decide if a typical user would naturally mention the platform when asking for this task.
3.  Your output MUST be a single line containing the classification **wrapped in `<coupling></coupling>` tags**. The value must be either `tightly_coupled` or `generic_concept`.

# EXAMPLE 1
## Platform: "GitHub"
## Tool Name: "create_pull_request"
## Tool Description: "Creates a new pull request to merge changes from one branch to another."
## Your Output:
<coupling>tightly_coupled</coupling>

# EXAMPLE 2
## Platform: "Google Drive"
## Tool Name: "create_document"
## Tool Description: "Creates a new blank document in the user's drive."
## Your Output:
<coupling>generic_concept</coupling>

# EXAMPLE 3
## Platform: "Stripe"
## Tool Name: "create_invoice"
## Tool Description: "Generates a new invoice for a customer."
## Your Output:
<coupling>generic_concept</coupling>

---

# YOUR TASK
## Platform: "{platform_name}"
## Tool Name: "{tool_name}"
## Tool Description: "{tool_description}"

## Your Output:
\end{lstlisting}

\noindent\textbf{Level 2 Query Generation Prompt}
\label{app:prompts_l2_query}

\begin{lstlisting}[
    basicstyle=\ttfamily\tiny,
    breaklines=true
]
# ROLE & GOAL
You are an expert user of various software, commanding an AI assistant. Your task is to write 2 natural-sounding user queries. The queries should reflect how a real human would ask for a task, based on how tightly the task is tied to its platform.

# CONTEXT FOR YOUR TASK
- **Platform Name**: "{platform_name}"
- **Tool Name**: "{tool_name}"
- **Tool Description**: "{tool_description}"
- **Tool Inputs**:
{formatted_schema}

# CORE INSTRUCTION
- **Platform Mention Rule**: {platform_mention_rule}

# GENERAL INSTRUCTIONS
1.  **Follow the Platform Mention Rule**: This is the most important instruction. Adhere strictly to whether you should or should not mention the platform name.
2.  **Embody the User**: Your tone must be that of a user giving a command.
3.  **AVOID META-LANGUAGE**: Do NOT refer to the tool itself (e.g., "use the tool").
4.  **Be Specific and Actionable**: Incorporate realistic example values for the parameters listed in "Tool Inputs".
5.  **Format**: Wrap each query in `<query></query>` tags.

# EXAMPLE 1: Tightly Coupled
## Platform Mention Rule: "You MUST mention the platform 'GitHub' because the function is iconic to it."
## Context: Tool is 'create_pull_request' on GitHub.
## Correct Output:
<query>Create a pull request on GitHub to merge the 'feature-x' branch into 'main'.</query>
<query>I need to open a new GitHub pull request for my latest changes.</query>

# EXAMPLE 2: Generic Concept
## Platform Mention Rule: "You MUST NOT mention the platform 'Google Drive'. The function is a generic concept."
## Context: Tool is 'create_document' on Google Drive.
## Correct Output:
<query>Create a new document for me titled 'Meeting Notes Q3'.</query>
<query>I need to start a new doc.</query>
---

# YOUR TASK
Now, using the context and the CORE INSTRUCTION provided at the top, generate 2 user queries.

# YOUR OUTPUT
\end{lstlisting}

\noindent\textbf{Level 2 Verification Prompt}
\label{app:prompts_l2_verification}

\begin{lstlisting}[
    basicstyle=\ttfamily\tiny,
    breaklines=true
]
# ROLE & GOAL
You are a highly discerning AI Routing Analyst. Your task is to verify if a generated user query is a high-quality, natural-sounding training example for a specific tool, following a given rule.

# CONTEXT
- **Tool's Platform**: "{platform_name}"
- **Tool Name**: "{tool_name}"
- **Tool Description**: "{tool_description}"

# VALIDATION CRITERIA
You must check the query against ALL of the following rules. If ANY rule is violated, the result is `false`.

1.  **Platform Mention Rule Adherence**: The query must strictly follow this rule: **{platform_mention_rule}**
2.  **Logical Routing Match**: Is the tool a direct and sensible way to fulfill the user's request?
3.  **Self-Contained & Unambiguous**: Is the query understandable on its own? It must not use vague references like "this code" unless a concrete example is embedded.
4.  **No Meta-Language**: Does the query sound like a real user? It must not refer to the tool itself (e.g., "use this tool", "run the function").

# YOUR TASK
Analyze the query below based on all the criteria, especially the Platform Mention Rule.

## Query to Verify: "{query}"

# YOUR RESPONSE
Structure your output as follows:
- First, provide the boolean judgement (`true` or `false`) **wrapped in `<is_match></is_match>` tags**.
- Second, on a new line, provide a one-sentence explanation for your decision, specifically mentioning which rule was passed or failed.
\end{lstlisting}

\noindent\textbf{Level 2 Bottom-Up Verification Prompt}
\label{app:prompts_l2_bottomup}

\begin{lstlisting}[
    basicstyle=\ttfamily\tiny,
    breaklines=true
]
Analyze these two tools and determine if they achieve the same end goal with the same scope and depth.

Tool A: {first_tool['tool_name']} ({first_tool['server_name']})
Description: {first_tool['description']}

Tool B: {second_tool['tool_name']} ({second_tool['server_name']})  
Description: {second_tool['description']}

Two tools are functionally equivalent if they achieve the same END RESULT with similar scope:
- Focus on WHAT they accomplish AND the breadth/depth of that accomplishment
- Consider the user's intent and the comprehensiveness of the output
- Different implementation methods are acceptable (different servers, models, APIs)
- But the scope, depth, and purpose of the end result should be equivalent

Answer "YES" if they produce the same type AND scope of result for users, "NO" if they have different scope, depth, or purpose.

Answer:"""
\end{lstlisting}

\noindent\textbf{Level 2 Round-Trip Verification Prompt}
\label{app:prompts_l2_roundtrip}

\begin{lstlisting}[
    basicstyle=\ttfamily\tiny,
    breaklines=true
]
# ROLE & GOAL
You are a highly intelligent AI routing engine. Your task is to analyze user queries and determine which of the provided candidate tools can fully and accurately resolve the request.

# CONTEXT
## User Query:
{query_list_text}

## Candidate Tools:
Here is a list of tools that might be able to help. Each tool is identified by a unique `tool_id`.

{candidate_tools_text}

# INSTRUCTIONS
1.  Read the user query to understand the core intent and requirements.
2.  For each candidate tool, evaluate if its function directly matches this core intent.
3.  A tool is a match ONLY IF its core function directly corresponds to the user's request. Do not select tools that are only partially related.
4.  Your final output MUST be a list of `tool_id`s for the tools you have selected.
5.  Each `tool_id` must be on a new line.
6.  The entire list of selected `tool_id`s MUST be wrapped in `<selected_tools>` and `</selected_tools>` tags.
7.  If you determine that NONE of the candidate tools are a suitable match, you must return an empty tag pair, like this: `<selected_tools></selected_tools>`.

# EXAMPLE
## User Query: "Create a new git repository for my project 'alpha'."
## Candidate Tools:
... (list of tools) ...
## Your Output (Example):
<selected_tools>
GitHub MCP Server::create_repository
GitLab MCP Server::create_project
</selected_tools>
\end{lstlisting}

\noindent\textbf{Level 3 Schema Inference Prompt}
\label{app:prompts_l3_infer_schema}

\begin{lstlisting}[
    basicstyle=\ttfamily\tiny,
    breaklines=true
]
Infer a plausible JSON output schema for the following tool:

Tool Name: {tool_name}
Description: {tool_description}
Input Schema: {input_schema}

Requirements:
1. Generate a realistic JSON schema that matches the tool's functionality
2. Include appropriate data types and structure
3. Make the schema specific but not overly complex

Output Schema:
{
  "type": "object",
  "properties": {
    // Your inferred properties here
  }
}
\end{lstlisting}

\noindent\textbf{Level 3 Logical Verification Prompt}
\label{app:prompts_l3_logical_verif}

\begin{lstlisting}[
    basicstyle=\ttfamily\tiny,
    breaklines=true
]
Verify if the following tool chain represents a logical, realistic workflow:

Tool Chain: {tool_chain}
Input-Output Flow: {flow_description}

Check:
1. Is the workflow logically coherent?
2. Are the input-output mappings realistic?
3. Does the sequence make practical sense?
4. Would this workflow be useful to users?

VERIFICATION: [PASS/FAIL]
REASONING: [Detailed explanation of your decision]
\end{lstlisting}

\noindent\textbf{Level 3 Query and CoT Generation Prompt}
\label{app:prompts_l3_query_cot}

\begin{lstlisting}[
    basicstyle=\ttfamily\tiny,
    breaklines=true
]
Generate a high-level user query and Chain-of-Thought plan for the following tool chain:

Tool Chain: {tool_chain}
Workflow: {workflow_description}

Requirements:
1. Create a realistic user query that would require this workflow
2. Generate a step-by-step Chain-of-Thought plan
3. Make the query specific and actionable
4. Ensure the plan matches the tool sequence

User Query: [Your generated query]
Chain-of-Thought Plan:
1. [First step]
2. [Second step]
3. [Third step]
...
\end{lstlisting}

\noindent\textbf{Level 4 Task Ideation Prompt}
\label{app:prompts_l4_task_ideation}

\begin{lstlisting}[
    basicstyle=\ttfamily\tiny,
    breaklines=true
]
Determine if a coherent cross-server task can be composed using the following servers:

Servers: {server_list}
Categories: {category_list}

Requirements:
1. Can these servers work together in a logical workflow?
2. Is the task realistic and useful for users?
3. Does the task require multiple servers to complete?
4. Are the server capabilities complementary?

Decision: [FEASIBLE/NOT_FEASIBLE]
Reasoning: [Detailed explanation of your decision]
If feasible, suggest a high-level task concept.
\end{lstlisting}

\noindent\textbf{Level 4 Feasibility Check Prompt}
\label{app:prompts_l4_feasibility}

\begin{lstlisting}[
    basicstyle=\ttfamily\tiny,
    breaklines=true
]
# ROLE & GOAL
You are a creative product manager designing tasks for a powerful AI assistant. Your task is to determine if a given combination of real-world services can form a logical user workflow.

# CONTEXT: PROVIDED SERVICES
Here are the services available for the potential task:
{services_description}

# YOUR TASK: FEASIBILITY CHECK
First, analyze the provided services. Can you imagine a realistic, common user scenario that would require **combining ALL** of these services? The workflow must be logical.

Respond with your boolean judgement (`true` or `false`) **wrapped in `<is_feasible></is_feasible>` tags**, followed by a brief, one-sentence reason for your decision on a new line.

# EXAMPLE 1 (Feasible)
## Services: [A GitHub Server, A Slack Server]
## Your Output:
<is_feasible>true</is_feasible>
A user might want to get notifications on Slack for new issues created in their GitHub repository.

# EXAMPLE 2 (Not Feasible)
## Services: [A Weather Forecast Server, A File Encryption Server]
## Your Output:
<is_feasible>false</is_feasible>
There is no common, logical workflow that directly connects weather forecasting with file encryption.

---

# YOUR CHECK NOW
## Services: [As provided in the context above]
## Your Output:
\end{lstlisting}

\noindent\textbf{Level 4 Workflow Generation Prompt}
\label{app:prompts_l4_workflow_generation}

\begin{lstlisting}[
    basicstyle=\ttfamily\tiny,
    breaklines=true
]
# SCENARIO GENERATION
Based on the services and tools provided below, please generate a user scenario.

# AVAILABLE SERVICES AND TOOLS:
{available_tools_section}

# HIGH-QUALITY EXAMPLES
Here are examples of well-structured queries and their corresponding ground truth tools:

## Example 1:
{{"query": "Get detailed information about the \\"My Favorite Chair\\" object within a Blender scene and check the PolyHaven integration status.", "ground_truth_tools_count": 2, "ground_truth_tools": [{{"tool_id": "Tripo MCP Server::get_object_info", "server_name": "Tripo MCP Server", "tool_name": "get_object_info", "description": "Tool for get_object_info functionality provided by Tripo MCP Server", "dependencies": []}}, {{"tool_id": "Tripo MCP Server::get_polyhaven_status", "server_name": "Tripo MCP Server", "tool_name": "get_polyhaven_status", "description": "Tool for get_polyhaven_status functionality provided by Tripo MCP Server", "dependencies": []}}]}}

## Example 2:
{{"query": "Scan the content of the file \\"main.py\\" using Semgrep, then retrieve the detailed results, check the status, and finally, get a list of supported languages for security vulnerability detection.", "ground_truth_tools_count": 4, "ground_truth_tools": [{{"tool_id": "Semgrep MCP Server::start_scan_from_content", "server_name": "Semgrep MCP Server", "tool_name": "start_scan_from_content", "description": "Tool for start_scan_from_content functionality provided by Semgrep MCP Server", "dependencies": []}}, {{"tool_id": "Semgrep MCP Server::get_scan_results", "server_name": "Semgrep MCP Server", "tool_name": "get_scan_results", "description": "Tool for get_scan_results functionality provided by Semgrep MCP Server", "dependencies": [0]}}, {{"tool_id": "Semgrep MCP Server::get_scan_status", "server_name": "Semgrep MCP Server", "tool_name": "get_scan_status", "description": "Tool for get_scan_status functionality provided by Semgrep MCP Server", "dependencies": [0]}}, {{"tool_id": "Semgrep MCP Server::get_supported_languages", "server_name": "Semgrep MCP Server", "tool_name": "get_supported_languages", "description": "Tool for get_supported_languages functionality provided by Semgrep MCP Server", "dependencies": []}}]}}

# INSTRUCTIONS
Based on the services and tools provided above, create a scenario following these guidelines:

1. **Specific and Actionable Query**: Write a detailed user query that includes:
   - Specific output formats (PDF, PPT, CSV, image, etc.) when relevant
   - File names and paths when relevant (e.g., /root/pdf/report.pdf, /home/user/data.csv)
   - Clear deliverables and requirements
   - Concrete subjects or targets (e.g., specific companies, topics, data sources)

2. **Ground Truth Tools**: For each tool needed in the workflow:
   - Use the exact tool names from the available services listed above
   - Set tool_id as "ServerName::tool_name"
   - Copy the description directly from the service definition or use the pattern "Tool for [tool_name] functionality provided by [server_name]"
   - Add dependencies as array of indices (0-based) if one tool depends on output from another
   - Include only essential tools needed for the workflow

3. **Natural Integration**: Ensure the query requires tools from ALL provided services in a logical workflow that a real user might request.

4. **Output Format**: Return your response as a valid JSON object with the exact structure shown in the examples above.

## Your Scenario Generation:
\end{lstlisting}

\noindent\textbf{Level 4 Quality Control Prompt}
\label{app:prompts_l4_quality_control}

\begin{lstlisting}[
    basicstyle=\ttfamily\tiny,
    breaklines=true
]
You are a professional query quality assessment and improvement expert. Please evaluate whether the following query contains the necessary parameters for executing the required tools, and improve the query when needed.

Original Query:
{query}

Expected tools and their parameter requirements:
{tools_description}

EVALUATION CRITERIA:

1. SOLVABILITY (1-10):
   - 10: All required data is provided, tools perfectly match needs, clear success criteria
   - 8-9: Task is clearly solvable with the given tools, minor ambiguities acceptable
   - 6-7: Mostly solvable but some steps may be challenging or unclear
   - 4-5: Significant gaps in tool coverage or data requirements
   - 1-3: Task cannot be meaningfully completed with available tools

   Consider:
   - Are all necessary tools available?
   - Is all required data provided (no external dependencies)?
   - Can the agent achieve the stated goal with these tools based on the function and output of the tools?
   - Are success criteria clear and measurable?

2. UTILITY (1-10):
   - 10: Critical business/research value, addresses real-world problem perfectly
   - 8-9: Strong practical value, useful for decision-making or operations
   - 6-7: Moderate value, interesting but not critical
   - 4-5: Limited practical value, mostly academic exercise
   - 1-3: Trivial or artificial task with no real-world application

   Consider:
   - Does this address a real business or research need?
   - Would the results be actionable and valuable?
   - Is the complexity justified by the outcome?
   - Does it test meaningful agent capabilities?

Provide scores and brief feedback in JSON format:
{{
  "solvability_score": <number 1-10>,
  "utility_score": <number 1-10>,
  "solvability_feedback": "Brief explanation of solvability assessment",
  "utility_feedback": "Brief explanation of utility assessment"
}}
\end{lstlisting}

\noindent\textbf{Equal Function Set Verification Prompt}
\label{app:prompts_l4_efs_verification}

\begin{lstlisting}[
    basicstyle=\ttfamily\tiny,
    breaklines=true
]
Given this specific query: {query_text}

Compare these two tools in the context of this query:

Original Ground Truth Tool: {original_tool['tool_name']} ({original_tool['server_name']})
Description: {original_tool['description']}

Candidate Replacement Tool: {candidate_tool['tool_name']} ({candidate_tool['server_name']})  
Description: {candidate_tool['description']}

In the context of the given query, can the candidate tool accomplish the same specific task as the original tool with equivalent scope and effectiveness?

Consider:
- The specific requirements of this query
- Whether both tools can produce the same type and quality of output needed for this query
- The role this tool plays in the multi-step workflow
- The specific data types, formats, and parameters involved

Answer "YES" if the candidate can effectively replace the original tool for this specific query, "NO" if it cannot.

Answer:
\end{lstlisting}

\noindent\textbf{Level 5 Capability Mapping Prompt}
\label{app:prompts_l5_capability_mapping}

\begin{lstlisting}[
    basicstyle=\ttfamily\tiny,
    breaklines=true
]
# ROLE & GOAL
You are an AI system architect. Your task is to analyze a group of tool descriptions and create a concise, high-level capability label for their shared function (e.g., "Manage Git Repositories").

# TOOL DESCRIPTIONS:
- {descriptions_text}

# YOUR OUTPUT
Wrap the final label in <label></label> tags.
\end{lstlisting}

\noindent\textbf{Level 5 Proposer Prompt}
\label{app:prompts_l5_proposer}

\begin{lstlisting}[
    basicstyle=\ttfamily\tiny,
    breaklines=true
]
ROLE & GOAL
As a world-class product manager for a general-purpose AI assistant, your goal is to brainstorm a diverse list of common digital tasks users might want to perform.
INSTRUCTIONS
List 100 distinct categories of digital tasks.
Cover a broad range of areas like productivity, entertainment, information retrieval, and e-commerce.
Output the result as a JSON list of strings.
EXAMPLE
["Travel Planning", "E-commerce Shopping", "Food & Local Services"]
YOUR OUTPUT
Wrap the JSON list inside <universal_tasks></universal_tasks> tags.
\end{lstlisting}

\noindent\textbf{Level 5 Red Team Prompt}
\label{app:prompts_l5_redteam}

\begin{lstlisting}[
    basicstyle=\ttfamily\tiny,
    breaklines=true
]
# ROLE & GOAL
You are a pragmatic and skeptical senior engineer, NOT a creative writer. Your goal is to find *realistic and direct* flaws in the claim that a task is impossible.

# CONTEXT
A "Proposer" AI claims the task **"{task}"** is impossible.
You are ONLY allowed to use the following **most relevant capabilities** to challenge this claim:
- {relevant_caps_text}

# YOUR TASK & STRICT RULES
Analyze if a challenge is valid based on these rules:
1.  **Rule of Direct Relevance**: The capability's primary function must be directly applicable to the task. Do NOT stretch the meaning of a capability. (e.g., A 'hashing' tool cannot be used for 'booking IDs').
2.  **Rule of Realistic Workflow**: A proposed solution must represent a simple, logical workflow that a user would find genuinely helpful. Do not chain more than 2-3 capabilities in a speculative way.
3.  **Final Decision**: Based on these strict rules, is there a valid, practical challenge to the "impossible" claim?

# YOUR OUTPUT
Provide a JSON object with two keys: {{"is_challenge_valid": boolean, "reasoning": "If valid, describe the realistic workflow. If invalid, state 'No practical or direct solution found.'"}}.
Wrap the JSON object in <challenge></challenge> tags.
\end{lstlisting}

\noindent\textbf{Level 5 Persona Query Generation Prompt}
\label{app:prompts_l5_persona_query}

\begin{lstlisting}[
    basicstyle=\ttfamily\tiny,
    breaklines=true
]
# ROLE & GOAL
You are acting as **{persona}**. Your goal is to generate {CONFIG['CANDIDATE_QUERIES_PER_GAP'] // len(personas)} realistic user queries for a task category you know an AI assistant cannot handle.
# CONTEXT
The assistant CANNOT do: "{gap}".
# YOUR TASK
Generate specific, natural-sounding queries that fall into the unsupported category: "{gap}".
# YOUR OUTPUT
List each query on a new line, starting with '-'. Wrap the list in <queries></queries> tags.
\end{lstlisting}

\noindent\textbf{Level 5 Final Verification Prompt}
\label{app:prompts_l5_final_verification}

\begin{lstlisting}[
    basicstyle=\ttfamily\tiny,
    breaklines=true
]
# ROLE & GOAL
You are a meticulous system evaluator. Analyze if the user query can be solved by the candidate tools and provide a structured JSON output.

# USER QUERY: "{query}"
# CANDIDATE TOOLS:
{candidate_tools_text}
# YOUR TASK
Follow these reasoning steps and output a single JSON object:
1.  **Query Intent Analysis**: What is the user's primary, concrete goal?
2.  **Direct Solution Analysis**: Is there any tool specifically designed for this intent?
3.  **Partial Solution Analysis**: If no direct tool exists, could generic tools plausibly assist the user?
4.  **Final Verdict**: Choose one: 'Directly Solvable', 'Partially Solvable', 'Out-of-Scope'.

# YOUR OUTPUT
Wrap the JSON object in <judgement></judgement> tags.
\end{lstlisting}

\section{Orchestrator Prompts}
\label{app:prompts}

This section provides the full prompt templates used in ETOM for each orchestrator. 
Prompts are grouped by orchestrator, with common prompts collected under a shared section.

\subsection{Shared Prompts}

\paragraph{Classification Prompt}\mbox{}\\
\label{app:classification_prompt}
\begin{lstlisting}[basicstyle=\ttfamily\tiny]
CLASSIFICATION_PROMPT = """You are a query classifier. Your task is to determine if a user query requires external tools/actions or can be answered using your existing knowledge.

Use the ReAct framework to analyze the query step by step:

**Thought Process:**
1. **Analyze**: What is the user asking for?
2. **Consider**: Does this require real-time data, external actions, or access to systems/files?
3. **Evaluate**: Can I answer this with my training knowledge alone?
4. **Decide**: Tools needed or conversational response?

**Classification Rules:**

**TOOLS Required** when the query involves:
- Real-time or current data (weather, stock prices, news)
- File system operations (read, write, create, delete files)
- External services (send email, make API calls, web searches)
- System operations (execute commands, manage processes)
- Database operations (queries, updates)
- Communication actions (send messages, notifications)
- Data retrieval from specific sources
- Actions that modify external state

**CONVERSATIONAL Response** when the query involves:
- General knowledge questions
- Explanations of concepts, theories, or how things work
- Creative tasks (writing, brainstorming, jokes)
- Analysis or comparison of known information
- Mathematical calculations or logical reasoning
- Advice or recommendations based on general principles
- Historical facts or established information

**Examples:**

TOOLS:
- "Get the current weather in Tokyo" (real-time data)
- "Read my notes.txt file" (file system access)
- "Send an email to John about the meeting" (external action)
- "Search for the latest Python tutorials" (web search)
- "Download the sales report from the server" (external data retrieval)
- "Execute ls command in the terminal" (system operation)
- "Get the current Bitcoin price" (real-time financial data)

CONVERSATIONAL:
- "What is the capital of France?" (general knowledge)
- "Explain quantum physics" (concept explanation)
- "How does machine learning work?" (educational content)
- "Tell me a joke" (creative content)
- "What are the benefits of exercise?" (general advice)
- "Compare Python and JavaScript" (known information comparison)
- "Calculate 15% of 200" (mathematical operation)

**Analysis Process:**
Query: "{query}"

Thought: Let me analyze this query step by step.
- What is being asked? [Identify the core request]
- Does this require external data or actions? [Yes/No reasoning]
- Can I answer this with my existing knowledge? [Yes/No reasoning]
- Final decision: [TOOLS or CONVERSATIONAL]

Action: Provide only the classification result.

Respond with only: TOOLS or CONVERSATIONAL
"""
\end{lstlisting}

\paragraph{Tool Selection Prompt}\mbox{}\\
\label{app:tool_selection_prompt}
\begin{lstlisting}[basicstyle=\ttfamily\tiny]
TOOL_SELECTION_PROMPT = """You are an AI assistant. Your task is to select the most appropriate tool from the list below based on the user's original query. 

IMPORTANT:
- Your response MUST be a **single valid JSON object**.
- Do NOT include any text, explanations, markdown code blocks, or commentary outside the JSON.
- Do NOT add trailing commas, comments, or any non-JSON content.
- Use double quotes for all keys and string values.
- Follow the schema exactly as shown below. Do NOT add or remove keys.
- For "server", use the exact "Server Name" from the tool list, NOT the "Server Description".

Expected JSON schema:
{{
    "tool_name": "<selected_tool_name>",
    "server": "<selected_server_name>",
    "arguments_kv": [
        {{"key": "<argument_name>", "value_json": "<json_encoded_value>"}},
        {{"key": "<argument_name>", "value_json": "<json_encoded_value>"}}
    ],
    "reasoning": "<brief explanation>"
}}

Instructions:
- If a suitable tool exists, fill in "tool_name" with the exact tool name and "server" with the exact server name.
- For "arguments_kv", provide a list of key-value pairs where each value is JSON-encoded:
  * For strings: "value_json": "\\"text\\""
  * For numbers: "value_json": "42"
  * For booleans: "value_json": "true"
  * For objects: "value_json": "{{\\"key\\": \\"value\\"}}"
  * For arrays: "value_json": "[1, 2, 3]"
- If no arguments are needed, use an empty array: "arguments_kv": []
- If no tool is suitable, set "tool_name" and "server" to "no", use empty arguments_kv, and explain why in "reasoning".
- Do not include any text outside the JSON object.
- Do not add comments or formatting outside the JSON.

Original query: {query}

Available tools:
{tools_list}
"""
\end{lstlisting}

\paragraph{Query Decomposition Prompt}\mbox{}\\
\label{app:query_decomposition_prompt}
\begin{lstlisting}[basicstyle=\ttfamily\tiny]
QUERY_DECOMPOSITION_PROMPT = """You are an expert at breaking down a complex user query into a list of self-contained, actionable, and tool-callable sub-tasks.
Your response must be a JSON object with a single key "sub_queries", which contains a list of strings.

- Each sub-task in the list must be a direct command that a tool can execute.
- Do NOT create sub-tasks for asking purely informational questions like "what is..." or "how does...".
- Information about the method, format, or constraints (e.g., "using API X", "in JSON format") must be kept within the main actionable sub-task.
- If the user's query is already a single actionable command, return it as a single-item list.

---
**Example 1: Complex Query**
User Query: "What's the weather like in New York tomorrow, and can you also find me a top-rated Italian restaurant near Times Square?"
Your Response:
{{"sub_queries": ["Get the weather forecast for tomorrow in New York City", "Find a top-rated Italian restaurant near Times Square"]}}

---
**Example 2: Simple Actionable Query**
User Query: "Can you please calculate the NPV for my project?"
Your Response:
{{"sub_queries": ["Calculate the NPV for the project"]}}

---
**Example 3: Tricky Query with Method/Constraint (VERY IMPORTANT)**
User Query: "Can you help me create floor plan views in Autodesk Revit for levels 1 and 2 using the JSON-RPC 2.0 method?"
Your Response:
{{"sub_queries": ["Create floor plan views in Autodesk Revit for levels 1 and 2 using the JSON-RPC 2.0 method"]}}
---

Now, decompose the following user query.

**User Query:** "{user_question}"
**Your Response:**
"""
\end{lstlisting}

\subsection{ReAct Orchestrator}

\paragraph{Server Selection Prompt}\mbox{}\\
\label{app:react_server_prompt}
\begin{lstlisting}[basicstyle=\ttfamily\tiny]
SERVER_SELECTION_PROMPT = """You are an AI assistant that helps users by analyzing their requests and selecting the most relevant platforms or services (servers) to fulfill the user's needs.

Given:
- A user request: {query}
- A list of available servers, each with a name and a brief description

Your task:
1. Carefully read and understand the user's request.
2. Review the list of servers and their descriptions.
3. Select the top-{top_k} servers that are most relevant to the user's request.
4. Respond using ONLY the following format:

<server_selection>
["server_name_1", "server_name_2", ..., "server_name_{top_k}"]
</server_selection>

Your response should be a JSON array of server names, ordered by relevance (most relevant first).
Remember to ONLY provide the JSON array within the <server_selection> tags. DO NOT provide any additional explanation or commentary outside the tags.

User request: {query}

Available servers:
{server_list}
"""
\end{lstlisting}

\paragraph{Tool Selection from Servers Prompt}\mbox{}\\
\label{app:react_tool_selection_prompt}
\begin{lstlisting}[basicstyle=\ttfamily\tiny]
TOOL_SELECTION_FROM_SERVERS_PROMPT = """You are an AI assistant that helps users by analyzing their requests and selecting the single most relevant tool from a given set of servers to fulfill the user's needs.

Given:
- A user request: {query}
- A list of tools from selected servers, each with server name, tool name, and tool description

Your task:
1. Carefully read and understand the user's request.
2. Review all tools from the selected servers and their descriptions.
3. Select the ONE tool that is most relevant and appropriate for the user's request.
4. Respond using ONLY the following format:

<tool_selection>
"tool_name"
</tool_selection>

Your response should be a single tool name (as a string, not an array).
Remember to ONLY provide the tool name within the <tool_selection> tags. DO NOT provide any additional explanation or commentary outside the tags.

User request: {query}

Available tools from selected servers:
{tool_list}
"""
\end{lstlisting}

\paragraph{Needle Selection Prompt}\mbox{}\\
\label{app:react_needle_prompt}
\begin{lstlisting}[basicstyle=\ttfamily\tiny]
NEEDLE_SELECTION_PROMPT = """You are an AI assistant that helps users by analyzing their requests and identifying appropriate tools. Your task is to identify both the SERVER (platform/service domain) and the specific TOOL (operation type + target) that would best address the user's request.

When a user asks you to perform a task, you should:
1. Carefully read and understand the user's request
2. Identify the key requirements and intentions in the request
3. Determine the information of the server from user's request (SERVER)
4. Determine what specific tool from user's request (TOOL)
5. Respond using ONLY the following format:

<tool_assistant>
server: [brief description of the server/platform from user's request]
tool: [brief description of the specific tool from user's request]
</tool_assistant>

Your response should be concise but descriptive.
Remember to ONLY provide the server and tool descriptions within the <tool_assistant> tags. DO NOT provide any additional explanation or commentary outside the <tool_assistant> tags.

User request: {query}You are an AI assistant that helps users by analyzing their requests and identifying appropriate tools. Your task is to identify both the SERVER (platform/service domain) and the specific TOOL (operation type + target) that would best address the user's request.

When a user asks you to perform a task, you should:
1. Carefully read and understand the user's request
2. Identify the key requirements and intentions in the request
3. Determine the information of the server from user's request (SERVER)
4. Determine what specific tool from user's request (TOOL)
5. Respond using ONLY the following format:

<tool_assistant>
server: [brief description of the server/platform from user's request]
tool: [brief description of the specific tool from user's request]
</tool_assistant>

Your response should be concise but descriptive.
Remember to ONLY provide the server and tool descriptions within the <tool_assistant> tags. DO NOT provide any additional explanation or commentary outside the <tool_assistant> tags.

User request: {query}
"""
\end{lstlisting}

\paragraph{Planning Prompt}\mbox{}\\
\label{app:react_planning_prompt}
\begin{lstlisting}[basicstyle=\ttfamily\tiny]
PLANNING_PROMPT = """You are an AI assistant that helps users by breaking down complex tasks into smaller, actionable steps using available tools.

Given:
- Original user request: {original_query}
- Current context and previous actions: {context}
- Current step in the plan

Your task:
1. Analyze the current progress and what still needs to be done
2. Determine the next specific action needed to progress toward the goal
3. Decide if the task is complete or if more steps are needed
4. If the query does not require tools or cannot be solved with available context/tools, set status to "finish"
5. Respond using ONLY the following format:

<planning>
{{
    "status": "continue" or "finish",
    "next_action": "description of what needs to be done next (if status is continue)",
    "reasoning": "brief explanation of why this action is needed, why the task is complete, why no tools are needed, or why the problem cannot be solved"
}}
</planning>

Important: Set status to "finish" if:
- The task is complete
- The query does not require any tools (e.g., conversational questions, general information requests)
- The problem cannot be solved with current available tools or context

Remember to ONLY provide the JSON within the <planning> tags.

Original request: {original_query}
Current context: {context}
"""
\end{lstlisting}

\subsection{ToolShed Orchestrator}

\paragraph{Rewrite Prompt}\mbox{}\\
\label{app:toolshed_rewrite_prompt}
\begin{lstlisting}[basicstyle=\ttfamily\tiny]
REWRITE_PROMPT = """You are a query rewriting assistant. Your task is to rewrite and clean up the user query to make it clearer and more searchable.

Original Query: {query}

Please rewrite this query to:
1. Fix any typos or grammatical errors
2. Expand abbreviations and acronyms
3. Clarify unclear or ambiguous terms
4. Maintain the original intent and meaning
5. Make it more suitable for tool search

Return only the rewritten query without any explanation.

Rewritten Query:
"""
\end{lstlisting}

\paragraph{Query Expansion Prompt}\mbox{}\\
\label{app:toolshed_expansion_prompt}
\begin{lstlisting}[basicstyle=\ttfamily\tiny]
QUERY_EXPANSION_PROMPT = """You are an expert at converting user questions to {num_variations} sentence variations that target different keywords and nuanced approaches with the goal to embed this query in a vector database to retrieve relevant tools across various industries.

Your goal is to craft {num_variations} nuanced sentence variations that target different aspects of understanding or solving the query.
For example, one sentence could focus on a detailed aspect of the user query, while another is more broad to cover more ground when embedding these sentences to retrieve the most relevant tools for the user query.

Before you start, understand this from a practical standpoint: The user question can be matched to a range of tools or solutions within the system, and your crafted variations should optimize for breadth and specificity. Write out your approach and plan for tackling this, then provide the {num_variations} sentences you would craft for the user question.

Think through your approach step by step, be intelligent, take a deep breath.

USER QUESTION: {user_question}

YOUR APPROACH, REASONING, AND {num_variations} SENTENCES:
"""
\end{lstlisting}

\paragraph{Rerank Prompt}\mbox{}\\
\label{app:toolshed_rerank_prompt}
\begin{lstlisting}[basicstyle=\ttfamily\tiny]
RERANK_PROMPT = """OK here are the results:
USER QUESTION EMBEDDED AND RETRIEVED TOOLS:
{user_question_results}
{variation_results}
===================
Based on these results, rank the top {top_k} most relevant tools to solve the user question. Just return the {top_k} tool names for each relevant tool.

Return as a JSON list: ["tool_name_1", "tool_name_2", ...]
"""
\end{lstlisting}

\subsection{MCP-Zero Orchestrator}

\paragraph{Tool Transformation Prompt}\mbox{}\\
\label{app:mcpzero_tooltrans_prompt}
\begin{lstlisting}[basicstyle=\ttfamily\tiny]
TOOL_TRANSFORMATION_PROMPT = """You are an AI assistant that helps users by analyzing their requests and identifying appropriate tools. Your task is to identify both the SERVER (platform/service domain) and the specific TOOL (operation type + target) that would best address the user's request.

When a user asks you to perform a task, you should:
1. Carefully read and understand the user's request
2. Identify the key requirements and intentions in the request
3. Determine the information of the server from user's request (SERVER)
4. Determine what specific tool from user's request (TOOL)
5. Respond using ONLY the following format:

<tool_assistant>
server: [brief description of the server/platform from user's request]
tool: [brief description of the specific tool from user's request]
</tool_assistant>

Your response should be concise but descriptive.
Remember to ONLY provide the server and tool descriptions within the <tool_assistant> tags. DO NOT provide any additional explanation or commentary outside the <tool_assistant> tags.

User request: {query}
"""
\end{lstlisting}

\subsection{Hybrid Orchestrator}

\paragraph{Query Expansion Prompt}\mbox{}\\
\label{app:hybrid_expansion_prompt}
\begin{lstlisting}[basicstyle=\ttfamily\tiny]
QUERY_EXPANSION_PROMPT = """You are an expert at converting user questions to {num_variations} sentence variations that target different keywords and nuanced approaches with the goal to embed this query in a vector database to retrieve relevant tools across various industries.

Your goal is to craft {num_variations} nuanced sentence variations that target different aspects of understanding or solving the query.
For example, one sentence could focus on a detailed aspect of the user query, while another is more broad to cover more ground when embedding these sentences to retrieve the most relevant tools for the user query.

Before you start, understand this from a practical standpoint: The user question can be matched to a range of tools or solutions within the system, and your crafted variations should optimize for breadth and specificity. Write out your approach and plan for tackling this, then provide the {num_variations} sentences you would craft for the user question.

Think through your approach step by step, be intelligent, take a deep breath.

USER QUESTION: {user_question}

YOUR APPROACH, REASONING, AND {num_variations} SENTENCES:
"""
\end{lstlisting}

\paragraph{Rerank Prompt}\mbox{}\\
\label{app:hybrid_rerank_prompt}
\begin{lstlisting}[basicstyle=\ttfamily\tiny]
RERANK_PROMPT = """OK here are the results:
USER QUESTION EMBEDDED AND RETRIEVED TOOLS:
{user_question_results}
{variation_results}
===================
Based on these results, rank the top {top_k} most relevant tools to solve the user question. Just return the {top_k} tool names for each relevant tool.

Return as a JSON list: ["tool_name_1", "tool_name_2", ...]
"""
\end{lstlisting}

\subsection{Evaluation Prompts}
\label{app:evaluation_prompts}

\noindent\textbf{Consistency Validation Prompt}
\label{app:consistency_validation_prompt}

\begin{lstlisting}[basicstyle=\ttfamily\tiny]
CONSISTENCY_VALIDATION_PROMPT = """You are a quality assurance expert. Your task is to validate whether a generated query can be answered by the specified tools.

**Query:** {query}
**Tools:** {tool_list}

**Validation Criteria:**
1. **Semantic Compatibility**: Can the query be reasonably interpreted as requiring these tools?
2. **Functional Suitability**: Are the tools capable of addressing the query's requirements?
3. **Input Compatibility**: Can the tools handle the inputs implied by the query?
4. **Output Relevance**: Will the tools produce outputs relevant to the query?

**Output Format:**
Provide your validation result:

VALID: [YES/NO]
CONFIDENCE: [HIGH/MEDIUM/LOW]
REASONING: [Detailed explanation of your validation decision]
ISSUES: [Any specific issues or concerns identified]"""
\end{lstlisting}

\noindent\textbf{Quality Assessment Prompt}
\label{app:quality_assessment_prompt}

\begin{lstlisting}[basicstyle=\ttfamily\tiny]
QUALITY_ASSESSMENT_PROMPT = """You are a benchmark quality assessor. Your task is to evaluate the quality of a generated query-tool pair.

**Query:** {query}
**Tools:** {tool_list}
**Level:** {level}

**Quality Dimensions:**
1. **Clarity**: Is the query clear and unambiguous?
2. **Realism**: Does the query represent a realistic user request?
3. **Appropriateness**: Is the query appropriate for the specified level?
4. **Complexity**: Does the query have the right level of complexity?
5. **Tool Alignment**: Do the tools match the query requirements?

**Scoring Scale:**
- **Excellent (5)**: Meets all criteria exceptionally well
- **Good (4)**: Meets most criteria well with minor issues
- **Fair (3)**: Meets basic criteria but has some issues
- **Poor (2)**: Has significant issues but is salvageable
- **Unacceptable (1)**: Major problems, should be rejected

**Output Format:**
Provide your assessment:

OVERALL_SCORE: [1-5]
DIMENSION_SCORES: {
  "clarity": [1-5],
  "realism": [1-5], 
  "appropriateness": [1-5],
  "complexity": [1-5],
  "tool_alignment": [1-5]
}
REASONING: [Detailed explanation of your assessment]
RECOMMENDATIONS: [Any suggestions for improvement]"""
\end{lstlisting}